\documentclass[11pt]{article}
\usepackage[a4paper, total={6.8in, 8in}]{geometry}

\usepackage[utf8]{inputenc}
\usepackage{amsmath}
\usepackage{amsfonts}
\usepackage{amssymb}
\usepackage{cite}
\usepackage{graphicx}
\graphicspath{{Figs/}}
\def\bdoc{\begin{document}}             \def\edoc{\end{document}}
\def\babs{\begin{abstract}}             \def\eabs{\end{abstract}}
\def\bcc{\begin{center}}                \def\ecc{\end{center}}
\def\ben{\begin{enumerate}}             \def\een{\end{enumerate}}
\def\bit{\begin{itemize}}               \def\eit{\end{itemize}}
\def\bpic{\begin{picture}}              \def\epic{\end{picture}}
\def\barr{\begin{array}}                \def\earr{\end{array}}
\def\btabu{\begin{tabular}}             \def\etabu{\end{tabular}}
\def\bcc{\begin{center}}                \def\ecc{\end{center}}
\def\beq{\begin{equation}}              \def\eeq{\end{equation}}
\def\beqn{\begin{eqnarray}}             \def\eeqn{\end{eqnarray}}
\def\beqnx{\begin{eqnarray*}}           \def\eeqnx{\end{eqnarray*}}
\def\beqnn{\begin{eqnarray*}}			\def\eeqnn{\end{eqnarray*}}
\def\bfig{\begin{figure}}               \def\efig{\end{figure}}
\def\bver{\begin{verb}}					\def\ever{\end{verb}}
\def\bbib{\begin{thebibliography}{999}}	\def\ebib{\end{thebibliography}}

\def\blue#1{{\color{blue}#1}}
\def\red#1{{\color{red}#1}}
\def\green#1{{\color{green}#1}}

\def\wt{\widetilde}
\def\wh{\widehat}

\def\zerob{{\bm 0}}
\def\oneb{{\bm 1}}

\def\ab{{\bm a}}
\def\bb{{\bm b}}
\def\cb{{\bm c}}
\def\db{{\bm d}}
\def\eb{{\bm e}}
\def\fb{{\bm f}}
\def\gb{{\bm g}}
\def\hb{{\bm h}}
\def\ib{{\bm i}}
\def\jb{{\bm j}}
\def\kb{{\bm k}}
\def\lb{{\bm l}}
\def\mb{{\bm m}}
\def\nb{{\bm n}}
\def\ob{{\bm o}}
\def\pb{{\bm p}}
\def\qb{{\bm q}}
\def\rb{{\bm r}}
\def\sb{{\bm s}}
\def\tb{{\bm t}}
\def\ub{{\bm u}}
\def\vb{{\bm v}}
\def\wb{{\bm w}}
\def\xb{{\bm x}}
\def\yb{{\bm y}}
\def\zb{{\bm z}}

\def\Ab{{\bm A}}
\def\Bb{{\bm B}}
\def\Cb{{\bm C}}
\def\Db{{\bm D}}
\def\Eb{{\bm E}}
\def\Fb{{\bm F}}
\def\Gb{{\bm G}}
\def\Hb{{\bm H}}
\def\Ib{{\bm I}}
\def\Jb{{\bm J}}
\def\Kb{{\bm K}}
\def\Lb{{\bm L}}
\def\Mb{{\bm M}}
\def\Nb{{\bm N}}
\def\Ob{{\bm O}}
\def\Pb{{\bm P}}
\def\Qb{{\bm Q}}
\def\Rb{{\bm R}}
\def\Sb{{\bm S}}
\def\Tb{{\bm T}}
\def\Ub{{\bm U}}
\def\Vb{{\bm V}}
\def\Wb{{\bm W}}
\def\Xb{{\bm X}}
\def\Yb{{\bm Y}}
\def\Zb{{\bm Z}}

\def\alphab{\bm{\alpha}}
\def\betab{\bm{\beta}}
\def\deltab{\bm{\delta}}
\def\etab{\bm{\eta}}
\def\epsilonb{\bm{\epsilon}}
\def\gammab{\bm{\gamma}}
\def\omegab{\bm{\omega}}
\def\etab{\bm{\eta}}
\def\thetab{\bm{\theta}}
\def\xib{\bm{\xi}}
\def\lambdab{\bm{\lambda}}
\def\taub{\bm{\tau}}
\def\phib{\bm{\phi}}
\def\mub{\bm{\mu}}
\def\psib{\bm{\psi}}
\def\chib{\bm{\chi}}
\def\sigmab{\bm{\sigma}}

\def\Deltab{\bm{\Delta}}
\def\Lambdab{\bm{\Lambda}}
\def\Phib{\bm{\Phi}}
\def\Psib{\bm{\Psi}}
\def\Sigmab{\bm{\Sigma}}

\def\Ac{{\cal A}}
\def\Bc{{\cal B}}
\def\Cc{{\cal C}}
\def\Dc{{\cal D}}
\def\Ec{{\cal E}}
\def\Fc{{\cal F}}
\def\Gc{{\cal G}}
\def\Hc{{\cal H}}
\def\Ic{{\cal I}}
\def\Jc{{\cal J}}
\def\Kc{{\cal K}}
\def\Lc{{\cal L}}
\def\Mc{{\cal M}}
\def\Nc{{\cal N}}
\def\Oc{{\cal O}}
\def\Pc{{\cal P}}
\def\Qc{{\cal Q}}
\def\Rc{{\cal R}}
\def\Sc{{\cal S}}
\def\Tc{{\cal T}}
\def\Uc{{\cal U}}
\def\Vc{{\cal V}}
\def\Wc{{\cal W}}
\def\Xc{{\cal X}}
\def\Yc{{\cal Y}}
\def\Zc{{\cal Z}}

\def\d#1{\,\mbox{d} #1}
\def\disp#1{\displaystyle{#1}}

\def\iii{\int_{-\infty}^{+\infty}}
\def\izi{\int_{0}^{\infty}}
\def\izpi{\int_{0}^{\pi}}
\def\izdpi{\int_{0}^{2\pi}}
\def\intd{\int\kern-.8em\int}
\def\intt{\int\kern-.8em\int\kern-.8em\int}
\def\intg{\int\kern-1.1em\int}

\def\xh{\widehat{x}}
\def\thetah{\widehat{\theta}}
\def\betah{\widehat{\beta}}

\def\xbh{\widehat{\xb}}
\def\ybh{\widehat{\yb}}
\def\thetabh{\widehat{\thetab}}
\def\etabh{\widehat{\etab}}
\def\betabh{\widehat{\betab}}

\def\xbhk{\widehat{\xb}^{k}}
\def\thetahk{\widehat{\theta}^{k}}
\def\betahk{\widehat{\beta}^{k}}

\def\xbhkp{\widehat{\xb}^{k+1}}
\def\thetahkp{\widehat{\theta}^{k+1}}
\def\betahkp{\widehat{\beta}^{k+1}}

\def\thetabhk{\widehat{\thetab}^{k}}
\def\betabhk{\widehat{\betab}^{k}}

\def\thetabhkp{\widehat{\thetab}^{k+1}}
\def\betabhkp{\widehat{\betab}^{k+1}}

\def\thetamin{\theta_{\mbox{\tiny min}}}
\def\thetamax{\theta_{\mbox{\tiny max}}}
\def\betamin{\beta_{\mbox{\tiny min}}}
\def\betamax{\beta_{\mbox{\tiny max}}}

\def\ra{\rightarrow}
\def\la{\leftarrow}
\def\da{\downarrow}
\def\ua{\uparrow}

\def\Ra{\Rightarrow}
\def\La{\Leftarrow}
\def\Da{\Downarrow}
\def\Ua{\Uparrow}

\def\lra{\longrightarrow}
\def\lla{\longleftarrow}
\def\Lra{\Longrightarrow}
\def\Lla{\longleftarrow}

\def\lrarr{\leftrightarrow}
\def\Lrarr{\Leftrightarrow}
\def\udarr{\updownarrow}
\def\Uparr{\Updownarrow}

\def\expf#1{\exp\left[ {#1} \right]}

\def\argmax#1#2{\arg\max_{#1}\left\{ #2 \right\}}
\def\argmin#1#2{\arg\min_{#1}\left\{ #2 \right\}}

\def\Prob#1{\mbox{Pr}\left\{#1\right\}}
\def\var#1{\mbox{Var}\left\{#1\right\}}
\def\cov#1{\mbox{Cov}\left\{#1\right\}}
\def\corr#1{\mbox{Corr}\left\{#1\right\}}
\def\trace#1{\mbox{Tr}\left\{#1\right\}}
\def\rang#1{\mbox{rang}\left\{#1\right\}}
\def\det#1{\mbox{d\'et}\left\{#1\right\}}

\def\gbu{\underline{\gb}}
\def\fbu{\underline{\fb}}
\def\zbu{\underline{\zb}}
\def\epsilonbu{\underline{\epsilonb}}
\def\thetabu{\underline{\thetab}}

\def\ve{{v_{\epsilon}}}
\def\sigmae{{\sigma_{\epsilon}}}
\def\Sigmae{{\Sigmab_{\epsilonb}}}
\def\Sigmaf{{\Sigmab_{\fb}}}
\def\Sigmag{{\Sigmab_{\gb}}}

\def\fbuh{\widehat{\fbu}}
\def\zbuh{\widehat{\fbu}}
\def\thetabuh{\widehat{\thetabu}}

\def\fbh{\widehat{\fb}}
\def\Pbh{\widehat{\Pb}}
\def\Sigmabh{\widehat{\Sigmab}}
\def\Abh{\widehat{\Ab}}
\def\Bbh{\widehat{\Bb}}
\def\thetabh{\widehat{\thetab}}
\def\Rbe{\Rb_{\epsilon}}
\def\Rbeh{\widehat{\Rbe}}

\def\Rjk{{R_j}_k}
\def\fbik{{\fb_i}_k}
\def\fbjk{{\fb_j}_k}

\def\mik{{m_i}_k}
\def\sigmaik{{\sigma_i^2}_k}
\def\Sigmaik{{\Sigmab_i}_k}
\def\alphaik{{\alpha_i}_k}
\def\betaik{{\beta_i}_k}

\def\muik{{\mu_i}_k}
\def\vik{{v_i}_k}

\def\mjk{{m_j}_k}
\def\sigmajk{{\sigma_j^2}_k}
\def\Sigmajk{{\Sigmab_j}_k}
\def\alphajk{{\alpha_j}_k}
\def\betajk{{\beta_j}_k}

\def\mujk{{\mu_j}_k}
\def\vjk{{v_j}_k}
\def\njk{{n_j}_k}

\def\alphae{\alpha_{\epsilon}}
\def\betae{\beta_{\epsilon}}
\def\alphaez{\alpha_{\epsilon_0}}
\def\betaez{\beta_{\epsilon_0}}
\def\alphaei{\alpha_{\epsilon_i}}
\def\betaei{\beta_{\epsilon_i}}
\def\alphaeiz{\alpha_{\epsilon_{i_0}}}
\def\betaeiz{\beta_{\epsilon_{i-0}}}

\def\mbz{{{\mb}_z}}
\def\Sigmabz{{{\Sigmab}_z}}
\def\diag#1{\mbox{diag}\left[#1\right]}

\def\sigmah{\widehat{\sigma}}
\def\fh{\widehat{f}}
\def\sigmaz{\sigma_z}

\def\sumi{\sum_{i=1}^{M} }
\def\sumj{\sum_{j=1}^{N} }
\def\lambdah{\wh{\lambda}}
\def\lambdabh{\wh{\lambdab}}

\def\det#1{\hbox{det}\left(#1\right)}
\def\defined{\,\shortstack{$\triangle$\\ =}\,}

\def\zmat{\left[\matrix{0}\right]}
\def\det#1{\hbox{det}(#1)}
\def\th{\widehat{t}}
\def\xh{\widehat{x}}
\def\lambdah{\widehat{\lambda}}
\def\muh{\widehat{\mu}}
\def\sigmah{\widehat{\sigma}}
\def\rhoh{\widehat{\rho}}
\def\betah{\widehat{\beta}}
\def\alphah{\widehat{\alpha}}
\def\tauh{\widehat{\tau}}

\def\tbh{\widehat{\tb}}
\def\mubh{\widehat{\mub}}
\def\sigmabh{\widehat{\sigmab}}
\def\rhobh{\widehat{\rhob}}
\def\oneb{\mbox{\mathbold 1}}

\def\bm#1{\mbox{\boldmath $#1$}}
\def\zerob{{\mathbold 0}}
\def\oneb{{\mathbold 1}}
\def\intg{\int\kern-1.1em\int}
\def\expf#1{\exp\left[ {#1} \right]}

\def\gbu{\underline{\gb}}
\def\fbu{\underline{\fb}}
\def\zbu{\underline{\zb}}
\def\epsilonbu{\underline{\epsilonb}}
\def\uthetab{\underline{\thetab}}

\def\sigmae{{\sigma_{\epsilon}}}
\def\Sigmae{{\Sigma_{\epsilonb}}}
\def\Sigmabe{{\Sigmab_{\epsilonb}}}
\def\Sigmaf{{\Sigmab_{\fb}}}
\def\Sigmag{{\Sigmab_{\gb}}}

\def\fbuh{\widehat{\fbu}}
\def\zbuh{\widehat{\fbu}}
\def\uthetabh{\widehat{\uthetab}}

\def\fbh{\widehat{\fb}}
\def\Sigmabh{\widehat{\Sigmab}}
\def\Abh{\widehat{\Ab}}
\def\thetabh{\widehat{\thetab}}

\def\Rjk{{R_j}_k}
\def\fbik{{\fb_i}_k}
\def\fbjk{{\fb_j}_k}

\def\mik{{m_i}_k}
\def\sigmaik{{\sigma_i^2}_k}
\def\Sigmaik{{\Sigmab_i}_k}
\def\alphaik{{\alpha_i}_k}
\def\betaik{{\beta_i}_k}

\def\muik{{\mu_i}_k}
\def\vik{{v_i}_k}

\def\mjk{{m_j}_k}
\def\sigmajk{{\sigma_j^2}_k}
\def\Sigmajk{{\Sigmab_j}_k}
\def\alphajk{{\alpha_j}_k}
\def\betajk{{\beta_j}_k}

\def\mujk{{\mu_j}_k}
\def\vjk{{v_j}_k}
\def\njk{{n_j}_k}

\def\mbz{{{\mb}_z}}
\def\Sigmabz{{{\Sigmab}_z}}

\def\vect{\mbox{vect}}
\def\lra{\longrightarrow}

\def\gir{g_i(\rb)}
\def\fjr{f_j(\rb)}
\def\zjr{z_j(\rb)}
\def\epsilonir{\epsilon_i(\rb)}
\def\gbr{\gb(\rb)}
\def\fbr{\fb(\rb)}
\def\zbr{\zb(\rb)}
\def\epsilonbr{\epsilonb(\rb)}
\def\fbhr{\fbh(\rb)}
\def\Sigmabhr{\Sigmabh(\rb)}

\def\mbzr{\mb_{z(\rb)}}
\def\Sigmabzr{\Sigmab_{z(\rb)}}
\def\Sigmagbzr{{\Sigmab_g}_{z(\rb)}}

\def\mjzjr#1{{m_{#1}}_{z_{#1}(\rb)}}
\def\sigmajzjr#1{{\sigma_{#1}}_{z_{#1}(\rb)}}

\def\Beta{B}
\def\Betab{\bm{\Beta}}
\def\Betabe{\bm{\Beta}_{\epsilonb}}

\def\fh{\widehat{f}}
\def\sigmah{\widehat{\sigma}}
\def\sigmaei{\sigma_{\epsilon_i}^2}

\def\thetah{\widehat{\theta}}
\def\sigmah{\widehat{\sigma}}

\def\Ftxm{\widetilde{F}_{X}(x_-)}
\def\Ftxp{\widetilde{F}_{X}(x_+)}

\def\NU{{\cal V}}

\def\tFx{\widetilde{F}_{X}}

\def\Fx{F_{X}(x)}
\def\fx{f_{X}(x)}

\def\Fxv{F_{X}(x)}
\def\fxv{f_{X}(x)}

\def\Fxi{{F}_{X_i}(x_i)}
\def\fxi{{f}_{X_i}(x_i)}

\def\Gnu{G_{\NU}(\nu)}
\def\gnu{g_{\NU}(\nu)}

\def\Fnu{F_{\NU}(\nu)}
\def\fnu{f_{\NU}(\nu)}

\def\Finnu#1{F^{-1}_{\NU}(#1)}

\def\Gnua#1{G_{\NU}(#1)}
\def\Gnub#1{G_{\NU}^{-1}(#1)}
\def\Gnuh{G_{\NU}^{-1}(1/2)}

\def\Ftx{\widetilde{F}_{X}(x)}
\def\ftx{\widetilde{f}_{X}(x)}

\def\Ftxi{\widetilde{F}_{X_i}(x_i)}
\def\ftxi{\widetilde{f}_{X_i}(x_i)}

\def\Ftxv{\widetilde{F}_{X}(x)}
\def\FtX#1{\widetilde{F}_{X}(#1)}
\def\ftxv{\widetilde{f}_{X}(x)}

\def\fxnu{f_{X|\NU}(x|\nu)}
\def\fxNU{f_{X|\NU}(x|\NU)}
\def\FxNU{F_{X|\NU}(x|\NU)}

\def\fxNUtheta{f_{X|\NU,\theta}(x|\NU,\theta)}
\def\FxNUtheta{F_{X|\NU,\theta}(x|\NU,\theta)}

\def\Fxnu{F_{X|\NU}(x|\nu)}
\def\Fxnui#1{F_{X|\NU}(x|\nu_#1)}

\def\FxN#1{F_{X|\NU}(#1)}
\def\FxNU{F_{X|\NU}(x|\NU)}
\def\Fxnun#1{F_{X|\NU}(x|#1)}
\def\Fxinun#1{F_{X_i|\NU}(x_i|#1)}
\def\fxnu{f_{X|\NU}(x|\nu)}
\def\fxNU{f_{X|\NU}(x|\NU)}

\def\Fxvnu{F_{X|\NU}(x|\nu)}
\def\Fxvnun#1{F_{X|\NU}(x|#1)}
\def\FxvNU{F_{X|\NU}(x|\NU)}

\def\FXvnu{F_{X|\NU}}

\def\fxvnu{f_{X|\NU}(x|\nu)}
\def\fxvNU{f_{X|\NU}(x|\NU)}

\def\FXvNU#1{{F}_{X|\NU}(#1|\NU)}
\def\FXvn#1{{F}_{X|\NU}(x|#1)}

\def\Fxtheta{F_{X|\theta}(x|\theta)}
\def\fxtheta{f_{X|\theta}(x|\theta)}

\def\Fxvtheta{F_{X|\theta}(x|\theta)}
\def\fxvtheta{f_{X|\theta}(x|\theta)}

\def\Ftxtheta{\widetilde{F}_{X|\theta}(x|\theta)}
\def\ftxtheta{\widetilde{f}_{X|\theta}(x|\theta)}

\def\Ftxvtheta{\widetilde{F}_{X|\theta}(x|\theta)}
\def\ftxvtheta{\widetilde{f}_{X|\theta}(x|\theta)}

\def\Fxnutheta{F_{X|\NU,\theta}(x|\nu,\theta)}
\def\fxnutheta{f_{X|\NU,\theta}(x|\nu,\theta)}

\def\ftheta{f_{\Theta}(\theta)}
\def\fthetaxnuz{f_{\Theta|X,\nu_0}(\theta|x,\nu_0)}
\def\fthetax{f_{\Theta|X}(\theta|x)}

\def\Fxvnutheta{F_{X|\NU,\theta}(x|\nu,\theta)}
\def\FxvNUtheta{F_{X|\NU,\theta}(x|\NU,\theta)}
\def\fxvnutheta{f_{X|\NU,\theta}(x|\nu,\theta)}
\def\Ftxvnutheta{\widetilde{F}_{X|\theta}(x|\theta)}
\def\ftxvnutheta{\widetilde{f}_{X|\theta}(x|\theta)}

\def\FxvNUm{F_{X|\NU}(x_-|\NU)}
\def\FxvNUp{F_{X|\NU}(x_+|\NU)}

\def\Fxinu{F_{X_i|\NU}(x_i|\nu)}
\def\fxinu{f_{X_i|\NU}(x_i|\nu)}

\def\Ftxi{\widetilde{F}_{X_i}(x_i)}
\def\ftxi{\widetilde{f}_{X_i}(x_i)}

\def\fxitheta{f_{X_i|\theta}(x_i|\theta)}
\def\ftxitheta{\widetilde{f}_{X_i|\theta}(x_i|\theta)}

\def\thetah{\widehat{\theta}}
\def\thetat{\widetilde{\theta}}

\def\lnuxv{l_{\nu|X}(\nu|x)}

\def\Lnuxva#1{L_{\nu|X}(#1 | x)}
\def\lnuxva#1{l_{\nu|X}(#1 | x)}

\def\Prob#1{\mbox{P}\left(#1\right)}

\def\lthetaxvnuz{l(\theta)}

\def\lthetaxvnuz{l_{\theta|x,\nu_0}(\theta | x,\nu_0)}
\def\fxnuztheta{f_{X|\nu_0,\theta}(x | \nu_0,\theta)}
\def\fxinuztheta{f_{X|\nu_0,\theta}(x_i | \nu_0,\theta)}
\def\fxvtheta{f_{X|\theta}(x | \theta)}
\def\lthetaxv{l_{\theta|X}(\theta | x)}
\def\ltthetaxv{\widetilde{l}_{\theta|X}(\theta | x)}

\def\ltheta{l(\theta )}
\def\lttheta{\widetilde{l}(\theta)}
\def\Lnu{L(\nu )}
\def\Lnui#1{L(\nu_#1)}
\def\Linu{L_i(\nu )}
\def\Linu{L_i(\nu )}
\def\LNU{L(\NU )}
\def\Lin#1{L^{-1}(#1)}
\def\L#1{L(#1)}
\def\Li#1{L_i(#1)}
\def\FNU#1{F_\NU(#1)}
\def\FinNU#1{F^{-1}_\NU(#1)}

\def\ftthetax{\tilde{f}_{\Theta|X}(\theta|x)}
\def\ER{R}

\def\Ndd#1#2#3{{\cal N}\left( #1 ;~ #2, #3 \right)}
\def\Cdd#1#2#3{{\cal C}\left( #1 ;~ #2, #3 \right)}
\def\Sdd#1#2#3#4{{\cal S}\left( #1 ;~ #2, #3, #4 \right)}
\def\Edd#1#2{{\cal E}\left( #1 ;~ #2 \right)}
\def\DEdd#1#2{{\cal DE}\left( #1 ;~ #2 \right)}
\def\Gdd#1#2#3{{\cal G}\left( #1 ;~ #2, #3 \right)}
\def\IGdd#1#2#3{{\cal IG}\left( #1 ;~ #2, #3 \right)}

\def\Xwr{X(\omega,\rb)}
\def\xwr{x(\omega,\rb)}

\def\muzw{\mu_{z}(\omega)}
\def\muzwr{\mu_{z(\rb)}(\omega)}
\def\szw{\sigma_{z}^2(\omega)}
\def\szwr{\sigma_{z(\rb)}^2(\omega)}
\def\pzw{p_{z}(\omega)}
\def\pzwr{p_{z(\rb)}(\omega)}

\def\hszw{\widehat{\sigma_{z}^2}(\omega)}
\def\hmuzw{\widehat{\mu}_{z}(\omega)}
\def\hpzw{\widehat{p}_{z}(\omega)}

\def\muzwrl{\mu_{z(\rb)}(\omega-l)}

\def\pxwrcz{p\left(\xwr ~|~ z\right)}
\def\pxwr{p\left(\xwr\right)}

\def\Xbw{\Xb(\omega)}
\def\xbw{\xb(\omega)}
\def\uXb{\underline{\Xb}}
\def\uxb{\underline{\xb}}
\def\pxbw{p(\xbw)}
\def\puxb{p(\uxb)}
\def\pzb{P(\zb)}

\def\Szw{S_z(\omega)}
\def\mubw{\mub(\omega)}

\def\Srw{S_{\rb}(\omega)}
\def\Swr{S_{\omega}(\rb)}
\def\Sbw{\Sb(\omega)}
\def\sbw{\sb(\omega)}
\def\Sbr{\Sb(\rb)}
\def\mubw{\mub(\omega)}
\def\cov#1{\mbox{cov}[#1]}

\def\pdf{\emph{pdf}}
\def\pmf{\emph{pmf}}
\def\dpdx#1#2{\frac{\partial #1}{\partial #2}}

\def\REM#1{}

\def\XXwr{X(\omega,\rb)}
\def\xxwr{x(\omega,\rb)}

\def\Xwr{X_{\omega}(\rb)}
\def\xwr{x_{\omega}(\rb)}
\def\Xrw{X_{\rb}(\omega)}
\def\xrw{x_{\rb}(\omega)}

\def\Ywr{Y_{\omega}(\rb)}
\def\ywr{y_{\omega}(\rb)}
\def\Yrw{Y_{\rb}(\omega)}
\def\yrw{y_{\rb}(\omega)}

\def\Ewr{E_{\omega}(\rb)}
\def\ewr{\epsilon_{\omega}(\rb)}
\def\Erw{E_{\rb}(\omega)}
\def\erw{\epsilon_{\rb}(\omega)}

\def\Xbr{\Xb(\rb)}
\def\xbr{\xb(\rb)}
\def\Xbw{\Xb(\omega)}
\def\xbw{\xb(\omega)}

\def\uXb{\underline{\Xb}}
\def\uxb{\underline{\xb}}
\def\uXbz{\underline{\Xb}_z}
\def\uxbz{\underline{\xb}_z}

\def\uthetab{\underline{\thetab}}
\def\uthetabz{\underline{\thetab}_z}

\def\Ywr{Y_{\omega}(\rb)}
\def\ywr{y_{\omega}(\rb)}
\def\Yrw{Y_{\rb}(\omega)}
\def\yrw{y_{\rb}(\omega)}
\def\Ybr{\Yb(\rb)}
\def\ybr{\yb(\rb)}
\def\Ybw{\Yb(\omega)}
\def\ybw{\yb(\omega)}
\def\uYb{\underline{\Yb}}
\def\uyb{\underline{\yb}}
\def\uYbz{\underline{\Yb}_z}
\def\uybz{\underline{\yb}_z}

\def\Ewr{E_{\omega}(\rb)}
\def\ewr{\epsilon_{\omega}(\rb)}
\def\Erw{E_{\rb}(\omega)}
\def\erw{\epsilon_{\rb}(\omega)}
\def\Ebr{Eb(\rb)}
\def\ebr{\epsilonb(\rb)}
\def\Ebw{Eb(\omega)}
\def\ebw{\epsilonb(\omega)}
\def\uEb{\underline{\Eb}}

\def\Erwp{E_{\rb}(\omega')}
\def\Xrwp{X_{\rb}(\omega')}

\def\xwmr{x_{\omega-1}(\rb)}
\def\muzwm{\mu_{z}(\omega-1)}

\def\rhoz{\rho_z}
\def\sez{\sigma_{\eta_z}^2}

\def\xwrp{x_{\omega}(\rb')}
\def\xrwp{x_{\rb}(\omega')}
\def\xwpr{x_{\omega'}(\rb)}
\def\xwprp{x_{\omega'}(\rb')}
\def\Xwrp{x_{\omega}(\rb')}
\def\Xwrp{X_{\omega}(\rb')}
\def\Xwpr{X_{\omega'}(\rb)}
\def\Xwprp{X_{\omega'}(\rb')}

\def\sigmaed{\sigmae^2}
\def\sigmaew{\sigmae(\omega)}
\def\sigmaedw{\sigmaed(\omega)}

\def\Sigmabe{\Sigmab_{\epsilon}}
\def\sigmabh{\widehat{\Sigmab}}
\def\xbh{\widehat{\xb}}
\def\zbh{\widehat{\zb}}
\def\qbh{\widehat{\qb}}
\def\sbh{\widehat{\sb}}

\def\espx#1#2{\mbox{E}_{#1}\left\{#2\right\}}
\def\esp#1{\mbox{E}\left\{#1\right\}}
\def\d#1{\mbox{~d}#1}

\def\Tr#1{\mbox{Tr}\left[#1\right]}

\def\pmatrix#1{\left[\barr{ccccccccccccccc} #1 \earr\right]}
\def\Bf{\bm{B}}
\def\gbh{\widehat{\gb}}
\def\dbh{\widehat{\db}}
\def\Tr#1{\mbox{Tr}\left[#1\right]}
\def\Cov#1{\mbox{Cov}\left[#1\right]}
\def\iid{i.i.d.~}

\def\cro#1{\left[#1\right]}
\def\acc#1{\left\{#1\right\}}
\def\aprio{\emph{a priori~}}
\def\apost{\emph{a posteriori~}}
\def\hbh{\widehat{\hb}}
\def\qh{\widehat{q}}
\def\fbt{\widetilde{\fb}}
\def\thetabt{\widetilde{\thetab}}

\def\mubt{\widetilde{\mub}}
\def\Sigmabt{\widetilde{\Sigmab}}
\def\qt{\widetilde{q}}
\def\vt{\widetilde{v}}
\def\zt{\widetilde{z}}
\def\Dbt{\widetilde{\Db}}
\def\alphat{\widetilde{\alpha}}
\def\betat{\widetilde{\beta}}
\def\abt{\widetilde{\ab}}
\def\ajt{\widetilde{a}_j}
\def\taut{\widetilde{\tau}}
\def\mut{\widetilde{\mu}}
\def\Zbt{\widetilde{\Zb}}
\def\Abt{\widetilde{\Ab}}
\def\zbt{\widetilde{\zb}}

\def\bloca#1#2#3{
\begin{picture}(50,50)
  \put(32,42){\makebox(0,0){#1}}
  \put(25,50){\vector(0,-1){20}}
  \put(0,10){\framebox(50,20){#2}}
  \put(0,0){\makebox(50,0){#3}}  
\end{picture}
}

\def\ugb{\underline{\gb}}
\def\fbh{\widehat{\fb}}
\def\zbh{\widehat{\zb}}
\def\qbh{\widehat{\qb}}
\def\thetabh{\widehat{\thetab}}
\def\Pbh{\widehat{\Pb}}

\def\sigmae{\sigma_{\epsilon}}
\def\REM#1{}
\def\fh{\widehat{f}}
\def\zh{\widehat{z}}
\def\disp{\displaystyle}
\def\oneb{{\boldmath{1}}}
\def\zerob{{\boldmath{0}}}
\def\d#1{\; \mbox{d} #1}
\def\expf#1{\exp\left\{#1\right\}}
\def\esp#1{\mbox{E}\left\{#1\right\}}
\def\lra{\longrightarrow}

\def\bslide#1{\begin{frame}{\Large\mathbold #1}}
\def\eslide{\end{frame}}

\def\inversions#1#2#3{
\begin{picture}(100,40)
  \put(0,15){\makebox(15,0){#1}}
  \put(20,15){\vector(1,0){10}}
  \put(30,7){\framebox(40,15){#2}}
  \put(70,15){\vector(1,0){10}}
  \put(85,15){\makebox(15,0){#3}}
\end{picture}
}

\def\rfb{\red{\fb}}
\def\bgb{\blue{\gb}}
\def\rfbh{\red{\widehat{\fb}}}
\def\rfbt{\red{\widetilde{\fb}}}
\def\rzbt{\red{\widetilde{\zb}}}
\def\rqb{\red{\qb}}
\def\rqbh{\red{\widehat{\qb}}}
\def\rzb{\red{\zb}}
\def\rzbh{\red{\widehat{\zb}}}
\def\rthetab{\red{\thetab}}
\def\retab{\red{\etab}}
\def\rthetabh{\red{\widehat{\thetab}}}
\def\rthetabt{\red{\widetilde{\thetab}}}
\def\retabt{\red{\widetilde{\etab}}}
\def\ralphabt{\red{\widetilde{\alphab}}}

\def\ugb{\underline{\gb}}
\def\fbh{\widehat{\fb}}
\def\zbh{\widehat{\zb}}
\def\qbh{\widehat{\qb}}
\def\thetabh{\widehat{\thetab}}
\def\Pbh{\widehat{\Pb}}

\def\UP#1{\uppercase{#1}}
\def\sca#1{\uppercase{#1}}

\def\sigmae{\sigma_{\epsilon}}
\def\REM#1{}
\def\fh{\widehat{f}}
\def\zh{\widehat{z}}
\def\disp{\displaystyle}
\def\oneb{{\boldmath{1}}}
\def\zerob{{\boldmath{0}}}
\def\d#1{\; \mbox{d} #1}
\def\expf#1{\exp\left\{#1\right\}}
\def\esp#1{\mbox{E}\left\{#1\right\}}
\def\lra{\longrightarrow}

\def\rfb{\red{\fb}}
\def\rFb{\red{\Fb}}
\def\bgb{\blue{\gb}}
\def\rfbh{\red{\widehat{\fb}}}
\def\rqb{\red{\qb}}
\def\rqbh{\red{\widehat{\qb}}}
\def\rzb{\red{\zb}}
\def\rzbh{\red{\widehat{\zb}}}
\def\rthetab{\red{\thetab}}
\def\ralphab{\red{\alphab}}
\def\rbetab{\red{\betab}}
\def\rthetabh{\red{\widehat{\thetab}}}
\def\ralphabh{\red{\widehat{\alphaab}}}
\def\rSigmab{\red{\Sigmab}}
\def\rSigmabe{\red{\Sigmabe}}

\def\rz{\red{z}}
\def\rzh{\red{\widehat{z}}}

\def\rab{\red{\ab}}
\def\rabh{\red{\widehat{\ab}}}
\def\rAb{\red{\Ab}}
\def\rAbh{\red{\widehat{\Ab}}}
\def\rBb{\red{\Bb}}
\def\rBbh{\red{\widehat{\Bb}}}
\def\rPbh{\red{\widehat{\Pb}}}
\def\rub{\red{\ub}}
\def\rubh{\red{\widehat{\ub}}}

\def\rVb{\red{\Vb}}
\def\rVbh{\widehat{\rVb}}
\def\rVbt{\widetilde{\rVb}}

\def\rvb{\red{\vb}}
\def\rvbh{\widehat{\rvb}}
\def\rvbt{\widetilde{\rvb}}

\def\bfb{\blue{\fb}}
\def\bFb{\blue{\Fb}}

\def\rfi{\red{f}_i}
\def\rfj{\red{f}_j}
\def\ruj{\red{u}_j}
\def\raj{\red{a}_j}
\def\rzj{\red{z}_j}
\def\rvj{\red{v}_j}
\def\rtau{\red{\tau}}
\def\rtauj{\red{\tau_j}}
\def\rtaub{\red{\taub}}
\def\rtauh{\widehat{\rtau}}
\def\rtaujh{\widehat{\rtauj}}
\def\rtaubh{\widehat{\rtaub}}
\def\rve{\red{v_{\epsilon}}}
\def\rv{\red{v}}

\def\rfjh{\red{\widehat{f}}_j}
\def\rujh{\red{\widehat{u}}_j}
\def\rajh{\red{\widehat{a}}_j}
\def\rzjh{\red{\widehat{z}}_j}
\def\rvjh{\red{\widehat{v}}_j}

\def\ralpha{\red{\alpha}}
\def\rbeta{\red{\beta}}
\def\rtheta{\red{\theta}}

\def\rlambda{\red{\lambda}}
\def\ralphah{\widehat{\ralpha}}
\def\rbetah{\widehat{\rbeta}}
\def\rthetah{\widehat{\rtheta}}
\def\rlambdah{\widehat{\rlambda}}

\def\ubh{\widehat{\ub}}

\def\rve{\red{v_{\epsilon}}}
\def\alphae{\alpha_{\epsilon_0}}
\def\betae{\beta_{\epsilon_0}}
\def\alphajh{\widehat{\alpha}_j}
\def\betajh{\widehat{\beta}_j}
\def\alphaeh{\widehat{\alpha}_{\epsilon_0}}
\def\betaeh{\widehat{\beta}_{\epsilon_0}}

\def\rqb{\red{\qb}}
\def\rqbh{\red{\widehat{\qb}}}
\def\rzb{\red{\zb}}
\def\rZb{\red{\Zb}}
\def\rab{\red{\ab}}
\def\rAb{\red{\Ab}}
\def\rvb{\red{\vb}}
\def\rmb{\red{\mb}}
\def\rzbh{\red{\widehat{\zb}}}
\def\rvbh{\red{\widehat{\vb}}}
\def\rmbh{\red{\widehat{\mb}}}
\def\rthetab{\red{\thetab}}
\def\rthetabh{\red{\widehat{\thetab}}}
\def\rfbh{\widehat{\rfb}}
\def\rhb{\red{\hb}}
\def\rhbh{\widehat{\rhb}}
\def\rthetai{\red{\theta}_i}
\def\rthetab{\red{\thetab}}
\def\rfjm{\red{f}_{j-1}}
\def\rfim{\red{f}_{i-1}}

\def\muh{\widehat{\mu}}
\def\vh{\widehat{v}}

\def\rSigmabh{\red{\Sigmabh}}
\def\rVbh{\red{\Vbh}}
\def\Fbh{\red{\Fb}}
\def\rFbh{\red{\Fbh}}

\def\Sigmabeh{\widehat{\Sigmabe}}
\def\rSigmabeh{\widehat{\red{\Sigmabe}}}

\def\rf{\red{f}}
\def\bg{\blue{g}}
\def\rF{\red{F}}
\def\bG{\blue{G}}

\def\sumk{\sum_{k=1}^K}
\def\suml{\sum_{l=1}^L}
\def\sumn{\sum_{n=1}^N}
\def\sumj{\sum_{j=1}^J}
\def\sumi{\sum_{i=1}^I}

\def\nub{\bm{\nu}}
\def\rak{\red{a_k}}
\def\rbk{\red{b_k}}
\def\rck{\red{c_k}}
\def\rtk{\red{t_k}}
\def\rvk{\red{v_k}}
\def\rnuk{\red{\nu_k}}
\def\rnuz{\red{\nu_0}}
\def\rphik{\red{\phi_k}}
\def\rnub{\red{\nub}}
\def\rtb{\red{\tb}}
\def\rphib{\red{\phib}}
\def\rfn{\red{f_n}}
\def\rtn{\red{t_n}}
\def\rvn{\red{v_n}}

\def\rve{\red{v_\epsilon}}
\def\rvf{\red{v_f}}
\def\rveh{\red{\widehat{v}_\epsilon}}
\def\rvfh{\red{\widehat{v}_f}}

\def\rnu{\red{\nu}}
\def\rphi{\red{\phi}}
\def\rbl{\red{b_l}}
\def\rbb{\red{\bb}}

\def\modela{\tt\setlength{\unitlength}{0.92pt}
\begin{picture}(250,106)
\thinlines    \put(-5,105){$\Nc(t|\red{t_1,v_1})$}
              \put(35,100){\vector(1,0){25}}
              \put(60,95){\framebox(37,16){\red{$a_1$}}}
			  \multiput(77,72)(0,7){3}{\circle*{4}}
              \put(-5,65){$\Nc(t|\red{t_k,v_k})$}
              \put(35,59){\vector(1,0){25}}
              \put(60,51){\framebox(37,16){\red{$a_k$}}}
              \multiput(77,32)(0,7){3}{\circle*{4}}              
              \put(-5,23){$\Nc(t|\red{t_K,v_K})$}
              \put(60,10){\framebox(37,16){\red{$a_K$}}}
              \put(35,18){\vector(1,0){25}}

              \put(98,103){\vector(1,-1){38}}
              \put(98,59){\vector(1,0){39}}
              \put(98,17){\vector(1,1){38}}

              \put(150,60){\circle{24}}
              \put(146,57){$\Large+$}
              \put(165,75){$g_{\rthetab}(t)$}
              \put(164,60){\vector(1,0){33}}
              
              \put(212,85){$\epsilon(t)$}
              \put(210,95){\vector(0,-1){20}}
              \put(211,60){\circle{24}}
              \put(207,58){$\Large+$}
              
              \put(225,66){$g(t)$}
              \put(225,60){\vector(1,0){20}}
\end{picture}
}

\def\modelb{\tt\setlength{\unitlength}{0.92pt}
\begin{picture}(250,106)
\thinlines    \put(-5,105){$\delta(t-\red{t_1})$}
              \put(0,100){\vector(1,0){37}}
              \put(38,95){\framebox(60,16){$\red{a_1} \Nc(t|0,\red{v_1})$}}
			  \multiput(77,72)(0,7){3}{\circle*{4}}
              \put(-5,65){$\delta(t-\red{t_k})$}
              \put(0,59){\vector(1,0){37}}
              \put(38,51){\framebox(60,16){$\red{a_k} \Nc(t|0,\red{v_k})$}}
              \multiput(77,32)(0,7){3}{\circle*{4}}              
              \put(-5,23){$\delta(t-\red{t_K})$}
              \put(38,10){\framebox(60,16){$\red{a_K} \Nc(t|0,\red{v_K})$}}
              \put(0,18){\vector(1,0){37}}

              \put(98,103){\vector(1,-1){38}}
              \put(98,59){\vector(1,0){39}}
              \put(98,17){\vector(1,1){38}}

              \put(150,60){\circle{24}}
              \put(146,57){$\Large+$}
              \put(165,75){$g_{\rthetab}(t)$}
              \put(164,60){\vector(1,0){33}}
              
              \put(212,85){$\epsilon(t)$}
              \put(210,95){\vector(0,-1){20}}
              \put(211,60){\circle{24}}
              \put(207,58){$\Large+$}
              
              \put(225,66){$g(t)$}
              \put(225,60){\vector(1,0){20}}
\end{picture}
}

\def\modelc{\tt\setlength{\unitlength}{0.92pt}
\begin{picture}(250,106)
\thinlines    \put(-5,105){$\delta(t-\red{t_1})$}
              \put(0,100){\vector(1,0){37}}
              \put(38,95){\framebox(40,16){$\red{a_1}$}}
			  \multiput(57,72)(0,7){3}{\circle*{4}}
              \put(-5,65){$\delta(t-\red{t_k})$}
              \put(0,59){\vector(1,0){37}}
              \put(38,51){\framebox(40,16){$\red{a_k}$}}
              \multiput(57,32)(0,7){3}{\circle*{4}}              
              \put(-5,23){$\delta(t-\red{t_K})$}
              \put(38,10){\framebox(40,16){$\red{a_K}$}}
              \put(0,18){\vector(1,0){37}}

              \put(78,103){\vector(1,-1){38}}
              \put(78,59){\vector(1,0){39}}
              \put(78,17){\vector(1,1){38}}

              \put(130,60){\circle{24}}
              \put(126,57){$\Large+$}
              \put(144,60){\vector(1,0){10}}
              \put(154,52){\framebox(50,16){$\Nc(t|0,v)$}}
              
              \put(205,75){$g_{\rthetab}(t)$}
              \put(204,60){\vector(1,0){33}}
              
              \put(252,85){$\epsilon(t)$}
              \put(250,95){\vector(0,-1){20}}
              \put(251,60){\circle{24}}
              \put(247,58){$\Large+$}
              
              \put(265,66){$g(t)$}
              \put(265,60){\vector(1,0){20}}
\end{picture}
}

\def\modeld{\tt\setlength{\unitlength}{0.92pt}
\begin{picture}(250,106)
\thinlines    \put(20,70){$s(t)=\sum_k \rak \delta(t-\red{t_k})$}
              \put(105,60){\vector(1,0){50}}
              \put(154,52){\framebox(50,16){$\Nc(t|0,v)$}}
              
              \put(205,75){$g_{\rthetab}(t)$}
              \put(204,60){\vector(1,0){33}}
              
              \put(252,85){$\epsilon(t)$}
              \put(250,95){\vector(0,-1){20}}
              \put(251,60){\circle{24}}
              \put(247,58){$\Large+$}
              
              \put(265,66){$g(t)$}
              \put(265,60){\vector(1,0){20}}
              \put(0,20){\siga}
\end{picture}
}

\def\siga{\setlength{\unitlength}{0.46pt}
\begin{picture}(290,108)
\thinlines    \put(10,26){\vector(1,0){260}}
              \put(16,12){\vector(0,1){88}}
              \put(48,27){\vector(0,1){34}}
              \put(79,27){\vector(0,1){54}}
              \put(130,27){\vector(0,1){46}}
              \put(188,26){\vector(0,1){40}}
              \put(262,14){$t$}
              \put(182,16){$\red{t_K}$}
              \put(195,63){$\red{a_K}$}
              \put(125,15){$\red{t_k}$}
              \put(137,68){$\red{a_k}$}
              \put(75,14){$\red{t_2}$}
              \put(84,73){$\red{a_2}$}
              \put(43,15){$\red{t_1}$}
              \put(54,56){$\red{a_1}$}
              \multiput(150,46)(8,0){3}{\circle*{4}}
\end{picture}
}

\def\modele{\tt\setlength{\unitlength}{0.92pt}
\begin{picture}(250,106)
\thinlines    \put(20,70){$s(t)=\sum_n \rfn \delta(t-n\Delta t)$}
              \put(105,60){\vector(1,0){50}}
              \put(154,52){\framebox(50,16){$\Nc(t|0,v)$}}
              
              \put(205,75){$g_{\rfb}(t)$}
              \put(204,60){\vector(1,0){33}}
              
              \put(252,85){$\epsilon(t)$}
              \put(250,95){\vector(0,-1){20}}
              \put(251,60){\circle{24}}
              \put(247,58){$\Large+$}
              
              \put(265,66){$g(t)$}
              \put(265,60){\vector(1,0){20}}
              \put(0,20){\sigb}
\end{picture}
}

\def\sigb{\setlength{\unitlength}{0.46pt}
\begin{picture}(290,108)
\thinlines    \put(10,26){\vector(1,0){260}}
              \put(10,12){\vector(0,1){88}}
              
              \put(20,27){\vector(0,1){34}}
              \put(40,27){\vector(0,1){14}}
              \put(60,27){\vector(0,1){54}}
              \put(80,27){\vector(0,1){31}}
              \put(100,27){\vector(0,1){46}}
              \put(120,27){\vector(0,1){40}}
              \put(140,27){\vector(0,1){48}}
              \put(160,27){\vector(0,1){34}}
              \put(180,27){\vector(0,1){14}}
              \put(200,27){\vector(0,1){31}}
              \put(220,27){\vector(0,1){46}}
              \put(240,27){\vector(0,1){40}}
              
              \put(270,14){$t$}
              \put(15,10){$1$}
              \put(15,63){$\red{f_1}$}
              \put(135,10){$n$}
              \put(135,68){$\red{f_n}$}
              \put(235,10){$N$}
              \put(235,68){$\red{f_N}$}

\end{picture}
}

\def\sigbr{\setlength{\unitlength}{0.46pt}
\begin{picture}(290,108)
\thinlines    \put(10,26){\vector(1,0){260}}
              \put(10,12){\vector(0,1){88}}
              
              \put(20,27){\vector(0,1){34}}
              \put(40,27){\vector(0,1){14}}
              \put(60,27){\vector(0,1){54}}
              \put(80,27){\vector(0,1){31}}
              \put(100,27){\vector(0,1){46}}
              \put(120,27){\vector(0,1){40}}
              \put(140,27){\vector(0,1){48}}
              \put(160,27){\vector(0,1){34}}
              \put(180,27){\vector(0,1){14}}
              \put(200,27){\vector(0,1){31}}
              \put(220,27){\vector(0,1){46}}
              \put(240,27){\vector(0,1){40}}
              \put(270,14){$t$}
              \put(15,10){$1$}
              \put(15,63){$\red{f_1}$}
              \put(135,10){$n$}
              \put(135,68){$\red{f_n}$}
              \put(235,10){$N$}
              \put(235,68){$\red{f_N}$}

              \put(60,10){$\red{\Delta t}$}
\end{picture}
}

\def\modelf{\tt\setlength{\unitlength}{0.92pt}
\begin{picture}(250,106)
\thinlines    \put(20,70){$s(t)=\sum_n \rfn \delta(t-n\red{\Delta t})$}
              \put(105,60){\vector(1,0){50}}
              \put(154,52){\framebox(50,16){$\Nc(t|0,\red{v})$}}
              
              \put(205,75){$g_{\rfb,\rthetab}(t)$}
              \put(204,60){\vector(1,0){33}}
              
              \put(252,85){$\epsilon(t)$}
              \put(250,95){\vector(0,-1){20}}
              \put(251,60){\circle{24}}
              \put(247,58){$\Large+$}
              
              \put(265,66){$g(t)$}
              \put(265,60){\vector(1,0){20}}
              \put(0,20){\sigbr}
\end{picture}
}

\def\sig{\tt\setlength{\unitlength}{1.38pt}
\begin{picture}(290,108)
\thinlines    
              \put(10,26){\vector(1,0){260}}
              \put(48,27){\vector(0,1){34}}
              \put(16,12){\vector(0,1){88}}
              \put(79,27){\vector(0,1){54}}
              \put(102,27){\vector(0,1){31}}
              \put(130,27){\vector(0,1){46}}
              \put(188,26){\vector(0,1){40}}
              \put(262,14){$t$}
              \put(22,90){$g(t)$}
              \put(182,16){$t_K$}
              \put(195,63){$a_K$}
              \put(125,15){$t_k$}
              \put(137,68){$a_k$}
              \put(75,14){$t_2$}
              \put(84,73){$a_2$}
              \put(43,15){$t_1$}
              \put(54,56){$a_1$}
              \multiput(150,46)(8,0){3}{\circle*{4}}
\end{picture}
}

\def\decgb{\tt\setlength{\unitlength}{0.92pt}
\begin{picture}(363,122)
\thinlines    \put(248,66){\vector(1,0){14}}
              \put(215,57){\framebox(33,18){$h(t)$}}
              \put(288,66){\vector(1,0){22}}
              \put(270,64){$\large\Sigma$}
              \put(275,67){\circle{26}}
              \put(275,102){\vector(0,-1){20}}
              \put(280,94){$b(t)$}
              \put(296,75){$y(t)$}
              \multiput(100,35)(0,8){3}{\circle*{4}}
              \put(194,72){$x(t)$}
              \put(193,66){\vector(1,0){23}}
              \put(176,61){$\large\Sigma$}
              \put(181,65){\circle{26}}
              \put(80,92){\framebox(43,18){$a_1$}}
              \put(80,58){\framebox(43,18){$a_2$}}
              \put(79,10){\framebox(43,18){$a_K$}}
              \put(123,66){\vector(1,0){45}}
              \put(123,101){\vector(4,-3){44}}
              \put(122,18){\vector(1,1){44}}
              \put(55,101){\vector(1,0){23}}
              \put(55,66){\vector(1,0){23}}
              \put(55,19){\vector(1,0){23}}
              \put(10,104){$g_1(t-t_1)$}
              \put(10,69){$g_2(t-t_2)$}
              \put(10,22){$g_K(t-t_k)$}
\end{picture}
}

\def\decgc{\tt\setlength{\unitlength}{0.92pt}
\begin{picture}(287,68)
\thinlines    \put(36,20){\vector(1,0){23}}
              \put(101,20){\vector(1,0){23}}
              \put(167,20){\vector(1,0){16}}
              \put(124,12){\framebox(43,18){$h(t)$}}
              \put(210,20){\vector(1,0){23}}
              \put(192,20){$\large\Sigma$}
              \put(196,23){\circle{26}}
              \put(196,57){\vector(0,-1){20}}
              \put(200,49){$b(t)$}
              \put(218,30){$y(t)$}
              \put(15,20){$s(t)$}
              \put(101,32){$x(t)$}
              \put(59,11){\framebox(43,18){$g(t)$}}
\end{picture}
}

\def\sigd{
\begin{picture}(290,108)
\thinlines    \put(10,26){\vector(1,0){260}}
              \put(48,27){\vector(0,1){34}}
              \put(16,12){\vector(0,1){88}}
              \put(79,27){\vector(0,1){54}}
              \put(102,27){\vector(0,1){31}}
              \put(130,27){\vector(0,1){46}}
              \put(188,26){\vector(0,1){40}}
              \put(262,14){$t$}
              \put(22,90){$g(t)$}
              \put(182,16){$t_K$}
              \put(195,63){$a_K$}
              \put(125,15){$t_k$}
              \put(137,68){$a_k$}
              \put(75,14){$t_2$}
              \put(84,73){$a_2$}
              \put(43,15){$t_1$}
              \put(54,56){$a_1$}
              \multiput(150,46)(8,0){3}{\circle*{4}}
\end{picture}
}

\def\wt#1{\widetilde{#1}}
\def\Thetab{\bm{\Theta}}
\def\Thetabh{\widehat{\Thetab}}
\def\abh{\widehat{\ab}}
\def\kbt{\tilde{\kb}}
\def\kt{\tilde{k}}
\def\gammat{\tilde{\gamma}}
\def\Sigmat{\tilde{\Sigma}}
\def\etat{\tilde{\eta}}

\def\rHb{\red{\Hb}}
\def\rFb{\red{\Fb}}

\def\wt#1{\widetilde{#1}}
\def\Thetab{\bm{\Theta}}
\def\Thetabh{\widehat{\Thetab}}
\def\abh{\widehat{\ab}}
\def\kbt{\tilde{\kb}}
\def\kt{\tilde{k}}
\def\gammat{\tilde{\gamma}}
\def\Sigmat{\tilde{\Sigma}}
\def\etat{\tilde{\eta}}
\def\zetat{\tilde{\zeta}}
\def\Mt{\tilde{M}}
\def\Mh{\hat{M}}
\def\Nt{\tilde{N}}
\def\atk{{\tilde{a}_k}}
\def\ztk{{\tilde{z}_k}}

\def\rh{\red{h}}
\def\rvbe{\rvb_{\epsilon}}
\def\rvei{\rv_{\epsilon_i}}
\def\rVbe{\rVb_{\epsilon}}
\def\rVbeh{\red{\Vbh}_{\epsilon}}
\def\vbh{\hat{\vb}}
\def\rVbfh{\red{\Vbh}_f}
\def\rvbeh{\rvbh_{\epsilon}}
\def\rvbet{\rvbt_{\epsilon}}
\def\rvbfh{\rvbh_f}

\def\rvbf{\rvb_f}
\def\rvfj{\rv_{f_j}}
\def\rVbf{\red{\Vb}_f}

\def\alphaez{\alpha_{\epsilon_0}}
\def\betaez{\beta_{\epsilon_0}}
\def\alphafz{\alpha_{f_0}}
\def\betafz{\beta_{f_0}}

\def\wh#1{\widehat{#1}}
\def\disp#1{{\displaystyle #1}}

\def\ER{\mbox{I\kern-.25em R}}
\def\EC{\mbox{C\kern-.8em C}}
\def\EZ{\mbox{Z\kern-.55em Z}}
\def\EN{\mbox{N\kern-.8em N}}

\def\bm#1{\mbox{\boldmath $#1$}}
\def\sb{{\bm s}}
\def\ub{{\bm u}}
\def\xb{{\bm x}}
\def\yb{{\bm y}}
\def\db{{\bm{d}}}
\def\Ab{{\bm A}}
\def\Bb{{\bm B}}
\def\Fb{{\bm F}}
\def\Rb{{\bm R}}
\def\Sb{{\bm S}}
\def\Ub{{\bm U}}

\def\thetab{\bm{\theta}}
\def\phib{\bm{\phi}}
\def\lambdab{\bm{\lambda}}
\def\Lambdab{\bm{\Lambda}}
\def\Sigmab{\bm{\Sigma}}
\def\lra{\longrightarrow}
\def\intg{\int\kern-1.1em\int}
\def\intd{\int\kern-.8em\int}

\def\apriori{{\em a priori} }
\def\aposteriori{{\em a posteriori} }

\def\d#1{\,\mbox{d}#1}
\def\esp#1{\mbox{E}\left\{ #1 \right\}}
\def\espx#1#2{\mbox{E}_{#1}\left\{ #2 \right\}}
\def\expf#1{\exp\left[ {#1} \right]}
\def\dpdx#1#2{\frac{\partial #1}{\partial #2}}
\def\dpdxd#1#2{\frac{\partial^2 #1}{\partial #2^2}}

\def\ent#1{\hbox{H}\left[#1\right]}
\def\dent#1{\delta\hbox{H}\left[#1\right]}
\def\rent#1{\hbox{D}\left[#1\right]}
\def\kullback#1{\hbox{K}\left[#1\right]}
\def\pent#1{\hbox{PE}\left[#1\right]}
\def\mdiv#1{\hbox{J}\left[#1\right]}
\def\infmut#1{\hbox{I}\left[#1\right]}
\def\idisc#1{\hbox{ID}\left[#1\right]}
\def\j{\mbox{j}}

\def\pxtheta{p(x | \thetab)}
\def\pxthetas{p(x | \thetab^*)}
\def\pxthetasd{p(x | \thetab^*+\Delta\thetab)}

\def\tr#1{\hbox{tr}\left(#1\right)}
\def\det#1{\hbox{det}\left(#1\right)}

\def\ieeeSSC{IEEE Transactions on Systems Science and Cybernetics}
\def\ieeeSP{IEEE Transactions on Signal Processing}			
\def\ieeeP{Proceedings of the IEEE}					
\def\ieeeIT{IEEE Transactions on Information Theory}			
\def\TS{Traitement du Signal}						
\def\siamCO{SIAM Journal of Control}					
\def\AISM{Annals of Institute of Statistical Mathematics}		
\def\ieeeASSP{IEEE Transactions on Acoustics Speech and Signal Processing}
\def\ieeeAC{IEEE Transactions on Automatic and Control}			

\def\jan{janvier\xspace}
\def\feb{f\'evrier\xspace}
\def\mar{mars\xspace}
\def\apr{avril\xspace}
\def\may{mai\xspace}
\def\jun{juin\xspace}
\def\jul{juillet\xspace}
\def\aug{ao\^ut\xspace}
\def\sep{septembre\xspace}
\def\oct{octobre\xspace}
\def\nov{novembre\xspace}
\def\dec{d\'ecembre\xspace}
\def\Jan{January\xspace}	
\def\Feb{February\xspace}
\def\Mar{March\xspace}
\def\Apr{April\xspace}
\def\May{May\xspace}
\def\Jun{June\xspace}
\def\Jul{July\xspace}
\def\Aug{August\xspace}
\def\Sep{September\xspace}
\def\Oct{October\xspace}
\def\Nov{November\xspace}
\def\Dec{December\xspace}

\def\espq#1{\left<#1\right>_q}

\def\rZbe{\rZb_{\epsilon}}
\def\nubt{\widetilde{\nub}}

\def\rVbf{\rVb_{f}}
\def\rvbf{\rvb_{f}}
\def\rvfj{\red{v}_{f_j}}
\def\rvbhe{\rvbh_{\epsilon}}
\def\rvbhf{\rvbh_{f}}
\def\bsplit{\begin{split}}
\def\esplit{\end{split}}
\def\alphaei{\alpha_{\epsilon_i}}
\def\alphafj{\alpha_{f_j}}
\def\betaei{\beta_{\epsilon_i}}
\def\betafj{\beta_{f_j}}

\def\bgbh{\widehat{\bgb}}
\def\rfh{\widehat{\rf}}
\def\rFh{\widehat{\rF}}
\def\rc{\red{r}}
\def\rch{\widehat{\rc}}
\def\partialx{\frac{\partial }{\partial x}}
\def\partialy{\frac{\partial }{\partial y}}
\def\partialr{\frac{\partial }{\partial r}}
\def\partialxd{\frac{\partial^2 }{\partial x^2}}
\def\partialyd{\frac{\partial^2 }{\partial y^2}}
\def\partialrd{\frac{\partial^2 }{\partial r^2}}
\def\cosphi{\cos(\phi)}
\def\sinphi{\sin(\phi)}
\def\sumk{\sum_{k=1}^K}
\def\sumj{\sum_{j=1}^N}
\def\sumi{\sum_{i=1}^M}
\def\summ{\sum_{m=1}^M}
\def\sumn{\sum_{n=1}^N}
\def\summn{\sum_{m=1}^M\sum_{n=1}^N}

\def\df{\dot{f}}
\def\ddf{\ddot{f}}

\def\rf{\red{f}}
\def\rdf{\red{\df}}
\def\rddf{\red{\ddf}}

\def\rfb{\red{\fb}}
\def\rdfb{\red{\dot{\fb}}}
\def\rddfb{\red{\ddot{\fb}}}

\def\bdg{\blue{\dot{g}}}
\def\bddg{\blue{\ddot{g}}}

\def\bgb{\blue{\gb}}
\def\bdg{\blue{\dot{g}}}

\def\bdgb{\blue{\dot{\gb}}}
\def\bddgb{\blue{\ddot{\gb}}}

\def\epsilonbd{\dot{\epsilonb}}
\def\epsilonbdd{\ddot{\epsilonb}}

\def\rdfbh{\widehat{\rdfb}}
\def\rddfbh{\widehat{\rddfb}}

\def\rqh{\red{\widehat{q}}}
\def\rdfh{\red{\widehat{\dot{f}}}}
\def\rddfh{\red{\widehat{\ddot{f}}}}

\def\vek{v_{\epsilon_k}}
\def\rveih{\widehat{\rve}_i}
\def\rvfjh{\widehat{\rvf}_j}

\def\alphaxiz{{\alpha_{\xi_0}}}
\def\betaxiz{{\beta_{\xi_0}}}
\def\alphazz{{\alpha_{z_0}}}

\def\betazz{{\beta_{z_0}}}
\def\rVbz{{\rVb_z}}
\def\rvbz{{\rvb_z}}
\def\rvxi{{\rv_\xi}}
\def\rvzj{{\rv_{z_j}}}
\def\rvxih{{\widehat{\rv}_\xi}}
\def\rvbzh{{\widehat{\rvb}_z}}
\def\rvzjh{{\widehat{\rv}_{z_j}}}

\def\alphae{{\alpha_\epsilon}}

\def\alphaet{{\widetilde{\alpha}_\epsilon}}
\def\alphaft{{\widetilde{\alpha}_f}}
\def\betae{{\beta_\epsilon}}
\def\betaet{{\widetilde{\beta}_\epsilon}}
\def\betaft{{\widetilde{\beta}_f}}
\def\rfih{\hat{\rf}_i}
\def\rfbk{\rfb^{(k)}}
\def\rfbkp{\rfb^{(k+1)}}

\def\rmh{\hat{\red{m}}}

\def\rmk{{\red{m}_k}}
\def\rmkh{{\hat{\red{m}}_k}}
\def\rvk{{\red{v}_k}}
\def\rvkh{{\hat{\red{v}}_k}}
\def\rak{{\red{a}_k}}
\def\rakh{{\hat{\red{a}}_k}}

\def\rvh{\hat{\red{v}}}
\def\rveh{\hat{\rve}}

\def\rmt{\tilde{\red{m}}}
\def\rvt{\tilde{\red{v}}}
\def\rvet{\tilde{\rve}}
\def\rveih{\red{\hat{v}}_{\epsilon_i}}

\def\rveb{\red{\vb_{\epsilon}}}
\def\rvebh{\red{\vbh_{\epsilon}}}

\def\blocgi{\begin{picture}(80,50)
  \put(60,40){\makebox(0,0){$\red{m}$}}
  \put(60,40){\circle{20}}
  \put(60,30){\vector(0,-1){20}}
  \put(60,0){\circle{20}}
  \put(60,0){\makebox(0,0){$\blue{g}_i$}}
  \put(20,0){\circle{20}}
  \put(20,0){\makebox(0,0){$\epsilon_i$}}
  \put(30,0){\vector(1,0){20}}
\end{picture}
}
\def\blocgfi{\begin{picture}(80,50)
  \put(60,40){\makebox(0,0){$\rf_i$}}
  \put(60,40){\circle{20}}
  \put(60,30){\vector(0,-1){20}}
  \put(60,0){\circle{20}}
  \put(60,0){\makebox(0,0){$\blue{g}_i$}}
  \put(20,0){\circle{20}}
  \put(20,0){\makebox(0,0){$\epsilon_i$}}
  \put(30,0){\vector(1,0){20}}
\end{picture}
}
\def\blocghfi{\begin{picture}(80,50)
  \put(60,40){\makebox(0,0){$\rf_i$}}
  \put(60,40){\circle{20}}
  \put(60,30){\vector(0,-1){20}}
  \put(65,20){\makebox(0,0){$h$}}
  \put(60,0){\circle{20}}
  \put(60,0){\makebox(0,0){$\blue{g}_i$}}
  \put(20,0){\circle{20}}
  \put(20,0){\makebox(0,0){$\epsilon_i$}}
  \put(30,0){\vector(1,0){20}}
\end{picture}
}
\def\blocgie{\begin{picture}(80,50)
  \put(60,40){\makebox(0,0){$\red{m}$}}
  \put(60,40){\circle{20}}
  \put(60,30){\vector(0,-1){20}}
  \put(60,0){\circle{20}}
  \put(60,0){\makebox(0,0){$\blue{g}_i$}}
  \put(30,0){\circle{20}}
  \put(30,0){\makebox(0,0){$\epsilon_i$}}
  \put(40,0){\vector(1,0){10}}
  \put(0,0){\circle{20}}
  \put(0,0){\makebox(0,0){$\rve$}}
  \put(10,0){\vector(1,0){10}}
\end{picture}
}
\def\blocgievei{\begin{picture}(80,50)
  \put(60,40){\makebox(0,0){$\red{m}$}}
  \put(60,40){\circle{20}}
  \put(60,30){\vector(0,-1){20}}
  \put(60,0){\circle{20}}
  \put(60,0){\makebox(0,0){$\blue{g}_i$}}
  \put(30,0){\circle{20}}
  \put(30,0){\makebox(0,0){$\epsilon_i$}}
  \put(40,0){\vector(1,0){10}}
  \put(0,0){\circle{20}}
  \put(0,0){\makebox(0,0){$\rvei$}}
  \put(10,0){\vector(1,0){10}}
\end{picture}
}
\def\blocfgie{\begin{picture}(80,100)
  \put(60,40){\makebox(0,0){$\rf_i$}}
  \put(60,40){\circle{20}}
  \put(60,30){\vector(0,-1){20}}
  \put(65,20){\makebox(0,0){$\blue{h}$}}
  
  \put(60,0){\circle{20}}
  \put(60,0){\makebox(0,0){$\blue{g}_i$}}
  \put(30,0){\circle{20}}
  \put(30,0){\makebox(0,0){$\epsilon_i$}}
  \put(40,0){\vector(1,0){10}}
  \put(0,0){\circle{20}}
  \put(0,0){\makebox(0,0){$\rve$}}
  \put(10,0){\vector(1,0){10}}
\end{picture}
}
\def\blocvfgie{\begin{picture}(80,100)
  \put(60,80){\makebox(0,0){$\rv_i$}}
  \put(60,80){\circle{20}}
  \put(60,70){\vector(0,-1){20}}
  
  \put(60,40){\makebox(0,0){$\rf_i$}}
  \put(60,40){\circle{20}}
  \put(60,30){\vector(0,-1){20}}
  \put(65,20){\makebox(0,0){$\blue{h}$}}
  
  \put(60,0){\circle{20}}
  \put(60,0){\makebox(0,0){$\blue{g}_i$}}
  \put(30,0){\circle{20}}
  \put(30,0){\makebox(0,0){$\epsilon_i$}}
  \put(40,0){\vector(1,0){10}}
  \put(0,0){\circle{20}}
  \put(0,0){\makebox(0,0){$\rve$}}
  \put(10,0){\vector(1,0){10}}
\end{picture}
}
\def\blocvfhgie{\begin{picture}(80,100)
  \put(60,80){\makebox(0,0){$\rv_i$}}
  \put(60,80){\circle{20}}
  \put(60,70){\vector(0,-1){20}}
  
  \put(60,40){\makebox(0,0){$\rf_i$}}
  \put(60,40){\circle{20}}
  \put(60,30){\vector(0,-1){20}}
  
  \put(30,30){\circle{20}}
  \put(30,30){\makebox(0,0){$\red{h}$}}
  \put(38,22){\vector(1,-1){15}}
  
  \put(60,0){\circle{20}}
  \put(60,0){\makebox(0,0){$\blue{g}_i$}}
  \put(30,0){\circle{20}}
  \put(30,0){\makebox(0,0){$\epsilon_i$}}
  \put(40,0){\vector(1,0){10}}
  \put(0,0){\circle{20}}
  \put(0,0){\makebox(0,0){$\rve$}}
  \put(10,0){\vector(1,0){10}}
\end{picture}
}
\def\blocvfhgiei{\begin{picture}(80,100)
  \put(60,80){\makebox(0,0){$\rv_i$}}
  \put(60,80){\circle{20}}
  \put(60,70){\vector(0,-1){20}}
  
  \put(60,40){\makebox(0,0){$\rf_i$}}
  \put(60,40){\circle{20}}
  \put(60,30){\vector(0,-1){20}}
  
  \put(30,30){\circle{20}}
  \put(30,30){\makebox(0,0){$\red{h}$}}
  \put(38,22){\vector(1,-1){15}}
  
  \put(60,0){\circle{20}}
  \put(60,0){\makebox(0,0){$\blue{g}_i$}}
  \put(30,0){\circle{20}}
  \put(30,0){\makebox(0,0){$\epsilon_i$}}
  \put(40,0){\vector(1,0){10}}
  \put(0,0){\circle{20}}
  \put(0,0){\makebox(0,0){$\rvei$}}
  \put(10,0){\vector(1,0){10}}
\end{picture}
}
\def\blocfgHe{\begin{picture}(80,100)
  \put(60,40){\makebox(0,0){$\rfb$}}
  \put(60,40){\circle{20}}
  \put(60,30){\vector(0,-1){20}}
  \put(65,20){\makebox(0,0){$\blue{\Hb}$}}
  
  \put(60,0){\circle{20}}
  \put(60,0){\makebox(0,0){$\bgb$}}
  \put(30,0){\circle{20}}
  \put(30,0){\makebox(0,0){$\epsilonb$}}
  \put(40,0){\vector(1,0){10}}
  \put(0,0){\circle{20}}
  \put(0,0){\makebox(0,0){$\rve$}}
  \put(10,0){\vector(1,0){10}}
\end{picture}
}
\def\blocvfgHe{\begin{picture}(80,100)
  \put(60,80){\makebox(0,0){$\rvb$}}
  \put(60,80){\circle{20}}
  \put(60,70){\vector(0,-1){20}}
  
  \put(60,40){\makebox(0,0){$\rfb$}}
  \put(60,40){\circle{20}}
  \put(60,30){\vector(0,-1){20}}
  \put(65,20){\makebox(0,0){$\bHb$}}
  
  \put(60,0){\circle{20}}
  \put(60,0){\makebox(0,0){$\bgb$}}
  \put(30,0){\circle{20}}
  \put(30,0){\makebox(0,0){$\epsilonb$}}
  \put(40,0){\vector(1,0){10}}
  \put(0,0){\circle{20}}
  \put(0,0){\makebox(0,0){$\rve$}}
  \put(10,0){\vector(1,0){10}}
\end{picture}
}
\def\blocvfgHei{\begin{picture}(80,100)
  \put(60,80){\makebox(0,0){$\rvb$}}
  \put(60,80){\circle{20}}
  \put(60,70){\vector(0,-1){20}}
  
  \put(60,40){\makebox(0,0){$\rfb$}}
  \put(60,40){\circle{20}}
  \put(60,30){\vector(0,-1){20}}
  \put(65,20){\makebox(0,0){$\bHb$}}
  
  \put(60,0){\circle{20}}
  \put(60,0){\makebox(0,0){$\bgb$}}
  \put(30,0){\circle{20}}
  \put(30,0){\makebox(0,0){$\epsilonb$}}
  \put(40,0){\vector(1,0){10}}
  \put(0,0){\circle{20}}
  \put(0,0){\makebox(0,0){$\rvei$}}
  \put(10,0){\vector(1,0){10}}
\end{picture}
}
\def\blocvfHge{\begin{picture}(80,100)
  \put(60,80){\makebox(0,0){$\rvb$}}
  \put(60,80){\circle{20}}
  \put(60,70){\vector(0,-1){20}}
  
  \put(60,40){\makebox(0,0){$\rfb$}}
  \put(60,40){\circle{20}}
  \put(60,30){\vector(0,-1){20}}
  
  \put(30,30){\circle{20}}
  \put(30,30){\makebox(0,0){$\rHb$}}
  \put(38,22){\vector(1,-1){15}}
  
  \put(60,0){\circle{20}}
  \put(60,0){\makebox(0,0){$\bgb$}}
  \put(30,0){\circle{20}}
  \put(30,0){\makebox(0,0){$\epsilonb$}}
  \put(40,0){\vector(1,0){10}}
  \put(0,0){\circle{20}}
  \put(0,0){\makebox(0,0){$\rve$}}
  \put(10,0){\vector(1,0){10}}
\end{picture}
}
\def\blocvfHgei{\begin{picture}(80,100)
  \put(60,80){\makebox(0,0){$\rvb$}}
  \put(60,80){\circle{20}}
  \put(60,70){\vector(0,-1){20}}
  
  \put(60,40){\makebox(0,0){$\rfb$}}
  \put(60,40){\circle{20}}
  \put(60,30){\vector(0,-1){20}}
  
  \put(30,30){\circle{20}}
  \put(30,30){\makebox(0,0){$\rHb$}}
  \put(38,22){\vector(1,-1){15}}
  
  \put(60,0){\circle{20}}
  \put(60,0){\makebox(0,0){$\bgb$}}
  \put(30,0){\circle{20}}
  \put(30,0){\makebox(0,0){$\epsilonb$}}
  \put(40,0){\vector(1,0){10}}
  \put(0,0){\circle{20}}
  \put(0,0){\makebox(0,0){$\rvei$}}
  \put(10,0){\vector(1,0){10}}
\end{picture}
}
\def\blocvhFge{\begin{picture}(80,100)
  \put(60,80){\makebox(0,0){$\rvb$}}
  \put(60,80){\circle{20}}
  \put(60,70){\vector(0,-1){20}}
  
  \put(60,40){\makebox(0,0){$\rhb$}}
  \put(60,40){\circle{20}}
  \put(60,30){\vector(0,-1){20}}
  
  \put(30,30){\circle{20}}
  \put(30,30){\makebox(0,0){$\bFb$}}
  \put(38,22){\vector(1,-1){15}}

  \put(60,0){\circle{20}}
  \put(60,0){\makebox(0,0){$\bgb$}}
  \put(30,0){\circle{20}}
  \put(30,0){\makebox(0,0){$\epsilonb$}}
  \put(40,0){\vector(1,0){10}}
  \put(0,0){\circle{20}}
  \put(0,0){\makebox(0,0){$\rve$}}
  \put(10,0){\vector(1,0){10}}
\end{picture}
}
\def\blocvhFgei{\begin{picture}(80,100)
  \put(60,80){\makebox(0,0){$\rvb$}}
  \put(60,80){\circle{20}}
  \put(60,70){\vector(0,-1){20}}
  
  \put(60,40){\makebox(0,0){$\rhb$}}
  \put(60,40){\circle{20}}
  \put(60,30){\vector(0,-1){20}}
  
  \put(30,30){\circle{20}}
  \put(30,30){\makebox(0,0){$\bFb$}}
  \put(38,22){\vector(1,-1){15}}

  \put(60,0){\circle{20}}
  \put(60,0){\makebox(0,0){$\bgb$}}
  \put(30,0){\circle{20}}
  \put(30,0){\makebox(0,0){$\epsilonb$}}
  \put(40,0){\vector(1,0){10}}
  \put(0,0){\circle{20}}
  \put(0,0){\makebox(0,0){$\rvei$}}
  \put(10,0){\vector(1,0){10}}
\end{picture}
}
\def\blochFge{\begin{picture}(80,40)
  \put(60,40){\makebox(0,0){$\rhb$}}
  \put(60,40){\circle{20}}
  \put(60,30){\vector(0,-1){20}}
  
  \put(30,30){\circle{20}}
  \put(30,30){\makebox(0,0){$\bFb$}}
  \put(38,22){\vector(1,-1){15}}

  \put(60,0){\circle{20}}
  \put(60,0){\makebox(0,0){$\bgb$}}
  \put(30,0){\circle{20}}
  \put(30,0){\makebox(0,0){$\epsilonb$}}
  \put(40,0){\vector(1,0){10}}
  \put(0,0){\circle{20}}
  \put(0,0){\makebox(0,0){$\rve$}}
  \put(10,0){\vector(1,0){10}}
\end{picture}
}
\def\blochFgei{\begin{picture}(80,40)
  \put(60,40){\makebox(0,0){$\rhb$}}
  \put(60,40){\circle{20}}
  \put(60,30){\vector(0,-1){20}}
  
  \put(30,30){\circle{20}}
  \put(30,30){\makebox(0,0){$\bFb$}}
  \put(38,22){\vector(1,-1){15}}

  \put(60,0){\circle{20}}
  \put(60,0){\makebox(0,0){$\bgb$}}
  \put(30,0){\circle{20}}
  \put(30,0){\makebox(0,0){$\epsilonb$}}
  \put(40,0){\vector(1,0){10}}
  \put(0,0){\circle{20}}
  \put(0,0){\makebox(0,0){$\rvei$}}
  \put(10,0){\vector(1,0){10}}
\end{picture}
}
\def\blocfHge{\begin{picture}(80,40)  
  \put(60,40){\makebox(0,0){$\bfb$}}
  \put(60,40){\circle{20}}
  \put(60,30){\vector(0,-1){20}}
  
  \put(30,30){\circle{20}}
  \put(30,30){\makebox(0,0){$\rHb$}}
  \put(38,22){\vector(1,-1){15}}
  
  \put(60,0){\circle{20}}
  \put(60,0){\makebox(0,0){$\bgb$}}
  \put(30,0){\circle{20}}
  \put(30,0){\makebox(0,0){$\epsilonb$}}
  \put(40,0){\vector(1,0){10}}
  \put(0,0){\circle{20}}
  \put(0,0){\makebox(0,0){$\rve$}}
  \put(10,0){\vector(1,0){10}}
\end{picture}
}
\def\blocfHgei{\begin{picture}(80,40)  
  \put(60,40){\makebox(0,0){$\bfb$}}
  \put(60,40){\circle{20}}
  \put(60,30){\vector(0,-1){20}}
  
  \put(30,30){\circle{20}}
  \put(30,30){\makebox(0,0){$\rHb$}}
  \put(38,22){\vector(1,-1){15}}
  
  \put(60,0){\circle{20}}
  \put(60,0){\makebox(0,0){$\bgb$}}
  \put(30,0){\circle{20}}
  \put(30,0){\makebox(0,0){$\epsilonb$}}
  \put(40,0){\vector(1,0){10}}
  \put(0,0){\circle{20}}
  \put(0,0){\makebox(0,0){$\rvei$}}
  \put(10,0){\vector(1,0){10}}
\end{picture}
}
\def\blocFgHe{\begin{picture}(80,40)
  \put(60,40){\makebox(0,0){$\bfb$}}
  \put(60,40){\circle{20}}
  \put(60,30){\vector(0,-1){20}}
  \put(65,20){\makebox(0,0){$\blue{\rHb}$}}
  
  \put(60,0){\circle{20}}
  \put(60,0){\makebox(0,0){$\bgb$}}
  \put(30,0){\circle{20}}
  \put(30,0){\makebox(0,0){$\epsilonb$}}
  \put(40,0){\vector(1,0){10}}
  \put(0,0){\circle{20}}
  \put(0,0){\makebox(0,0){$\rve$}}
  \put(10,0){\vector(1,0){10}}
\end{picture}
}

\def\ftrfi{\bg(r,\phi)}
\def\wth{\widehat{\tilde{h}}}
\def\varx#1#2{\mbox{Var}_{#1}\left\{#2\right\}}
\def\rvbxi{\rvb_{\xi}}
\def\rVbxi{\rVb_{\xi}}
\def\rvbxih{\rvbh_{\xi}}

\def\repsilon{\red{\epsilon}}

\def\vzj{v_{z_j}}
\def\vxij{v_{\xi_j}}
\def\vei{v_{\epsilon_i}}

\def\vzjh{\widehat{v}_{z_j}}
\def\vxijh{\widehat{v}_{\xi_j}}
\def\veih{\widehat{v}_{\epsilon_i}}

\def\vbeh{\widehat{\vb}_{\epsilon}}
\def\vbxih{\widehat{\vb}_{\xi}}
\def\vbzh{\widehat{\vb}_{z}}

\def\Vbeh{\widehat{\Vb}_{\epsilon}}
\def\Vbxih{\widehat{\Vb}_{\xi}}
\def\Vbzh{\widehat{\Vb}_{z}}

\def\Ybeh{\widehat{\Yb}_{\epsilon}}
\def\Ybxih{\widehat{\Yb}_{\xi}}
\def\Ybzh{\widehat{\Yb}_{z}}

\def\vf{v_f}
\def\vbe{\vb_{\epsilon}}
\def\vbz{\vb_z}
\def\vxi{v_{\xi}}
\def\vbxi{\vb_{\xi}}
\def\Vbe{\Vb_{\epsilon}}
\def\Vbxi{\Vb_{\xi}}
\def\Vbz{\Vb_z}

\def\half{\frac{1}{2}}

\def\dpdx#1#2{\frac{\partial {#1}}{\partial {#2}}}
\def\dpdxd#1#2{\frac{\partial^2 {#1}}{\partial {#2}^2}}
\def\dpdxdy#1#2#3{{{\partial ^2 {#1}}\over{\partial {#2} \partial {#3}}}}

\def\KL#1#2{\mbox{KL}\left[#1 : #2\right]}
\def\Covx#1#2{\mbox{Cov}_{#1}(#2)}
\def\braket#1#2{<{#1}(#2)>}

\def\rlambdab{\red{\lambdab}}
\def\bve{\blue{\ve}}
\def\ralphafjt{\red{\tilde{\alpha}_{f_j}}}
\def\rbetafjt{\red{\tilde{\beta}_{f_j}}}
\def\rSigmabt{\red{\tilde{\Sigmab}}}
\def\rfjt{\red{\tilde{f_j}}}
\def\zetab{\bm{\zeta}}
\def\bvxi{\blue{v_\xi}}
\def\bgamma{\blue{\gamma}}
\def\rgamma{\red{\gamma}}
\def\alphagamz{\alpha_{\gamma_0}}
\def\betagamz{\beta_{\gamma_0}}
\def\alphazetaz{\alpha_{\zeta_0}}
\def\betazetaz{\beta_{\zeta_0}}

\def\bvf{\blue{v_f}}
\def\bvzeta{\blue{v_\zeta}}
\def\rgb{\red{\gb}}
\def\rvxii{\red{v_{\xi_i}}}
\def\rvzeta{\red{v_{\zeta}}}

\usepackage{tikz,mathpazo}
\usetikzlibrary{shapes.geometric, arrows}
\usetikzlibrary{positioning,shapes,shadows,arrows}
\usetikzlibrary{mindmap,backgrounds}
\tikzstyle{startstop} = [rectangle, rounded corners, minimum width=1.5cm, minimum height=0.5cm,text centered, draw=black]
\tikzstyle{io} = [trapezium, trapezium left angle=70, trapezium right angle=110, minimum width=1.5cm, minimum height=0.6cm, text centered, draw=black]
\tikzstyle{process} = [rectangle, minimum width=1cm, minimum height=0.5cm, text centered, draw=black]
\tikzstyle{note} = [text centered]
\tikzstyle{decision} = [diamond, minimum width=1.5cm, minimum height=0.3cm, text centered, draw=black]
\tikzstyle{arrow} = [thick,->,>=stealth]

\def\blambda{\blue{\lambda}}
\def\blambdab{\blue{\lambdab}}

\def\rWb{\red{\Wb}}
\def\rWbh{\red{\widehat{\Wb}}}

\title{Deep Learning and Bayesian inference for Inverse Problems}

\author{Ali Mohammad-Djafari $^{1,2}$ orcid number:{0000-0003-0678-7759}, \\ 
Ning Chu$^{2,3}$, Li Wang$^{4}$, Liang Yu$^{5,6}$ 
\\ ~ \\ 
{\small \btabu{l}
$^1$ \quad International Science Consulting and Training (ISCT), 91440 Bures sur Yvette, France; djafari@ieee.org
\\
$^2$ \quad Zhejiang Shangfeng special blower company, Shaoxing 312352, China; chuning1983@sina.com
\\
$^3$ \quad Mechanical and Electrical Eng. Coll., Hainan Vocational Univ of Science and Tech. Haikou 571126, China
\\
$^4$ \quad Central South University, Changsha, China; li.wang.csu@csu.edu.cn
\\ 
$^5$ \quad Shanghai Jiao Tong Univ., Shaghai 200240, China; Liang.Yu@sjtu.edu.cn 
\\
$^6$ \quad Northwestern Polytechnical Univ., Xian 710072, China. 
\\ ~\\ 
Presented at MaxEnt 2023:   
International Workshop on Bayesian Inference and Maximum Entropy \\ 
Methods in Science and Engineering, 
Max Planck Institut, Garching, Germany, July 3-7, 2023.
\\ 
A modified version to appear in MaxEnt2023 Proceedings, published by MDPI. 
\etabu
}
}

\date{}

\begin{document}
\maketitle

\begin{abstract}
Inverse problems arise anywhere we have indirect measurement. As, in general they are ill-posed, to obtain satisfactory solutions for them needs prior knowledge. Classically, different regularization methods and Bayesian inference based methods have been proposed. As these methods need a great number of forward and backward computations, they become costly in computation, in particular, when the forward or generative models are complex and the evaluation of the likelihood becomes very costly. Using Deep Neural Network surrogate models and approximate computation can become very helpful. However, accounting for the uncertainties, we need first understand the Bayesian Deep Learning and then, we can see how we can use them for inverse problems. 
\\ 
In this work, we focus on NN, DL and more specifically the Bayesian DL particularly adapted for inverse problems. We first give details of Bayesian DL approximate computations with exponential families, then we will see how we can use them for inverse problems. 
We consider two cases: First the case where the forward operator is known and used as physics constraint, the second more general data driven DL methods.  
\\
{\bf keyword:}{Neural Network, Variational Bayesian inference, Bayesian Deep Learning (DL), Inverse problems, Physics based DL}
\end{abstract}

\maketitle

\section{Introduction}
Inverse problems arise almost everywhere in science and Engineering. In fact anytime and in any application, when we have indirect measurements, related to what we really want to measure through some mathematical relation, called Forward model. Then, we have to infer on the desired unknown from the observed data, using this forward model or a surrogate one. 
As, in general, many inverse problems are ill-posed, many methods for finding well-posed solutions for them are mainly based either on the regularization theory or the Bayesian inference. We may mention those, in particular, which are based on the optimization of a criterion with two parts: 
a data-model output matching criterion (likelihood part in the Bayesian) and a regularization term (prior model in the Bayesian). Different criteria for these two terms and a great number of standard and advanced optimization algorithms have been proposed and used with great success.  
When these two terms are distances, they can have a Bayesian Maximum A Posteriori (MAP) interpretation where these two terms correspond, respectively, to the likelihood and prior probability models. 
The Bayesian approach gives more flexibility in choosing these terms via the likelihood and the prior probability distributions. This flexibility goes much farther with the hierarchical models and appropriate hidden variables.  
\cite{5445046}. 
Also, the possibility of estimating the hyper parameters gives much more flexibility for semi-supervised methods.  

However, the full Bayesian computations can become very heavy computationally. In particular when the forward model is complex and the evaluation of the likelihood needs high computational cost. 
\cite{9234749}. 
In those cases using surrogate simpler models can become very helpful to reduce the computational costs, but then, we have to account for uncertainty quantification (UQ) of the obtained results \cite{bayesian2018zhu}. 
Neural Networks (NN) with their diversity such as Convolutional (CNN), Deep learning (DL), etc., have become tools as fast and low computational surrogate forward models for them. 

In the last decades, the Machine Learning (ML) methods and algorithms have gained great success in many tasks, such as classification, clustering, segmentation, object detection and many other area. There are many different structures of Neural Networks (NN), such feed-forward,  Convolutional NN (CNN), Deep NN, etc. 
\cite{review2017unser}.  
Using these methods directly for inverse problems, as intermediate pre-processing or as tools for doing fast approximate computation in different steps of regularization or Bayesian inference have also got success, but not as much as they could. 
Recently, the Physics-Informed Neural Networks have gained great success in many inverse problems, proposing interaction between the Bayesian formulation of forward models,  optimization algorithms and ML specific algorithms for intermediate hidden variables. 
These methods have become very helpful to obtain approximate practical solutions to inverse problems in real world applications 
\cite{9403414,8878159,9000801,8901171,8434321,physicsinformed2019chen,physicsinformed2019raissi}.  
 
In this paper, first, in Section 2, some mathematical notations for dealing with NN are given. In section 3, a detailed presentation of the Bayesian inference and approximate computation needed for BDL is given. Then, in section 4, we consider we focus on the NN and DL methods for inverse problems. First, we present same cases where we know the forward and its adjoint model. Then, we consider the case we may not have this knowledge and want to propose directly data driven DL methods.  
\cite{1029279,AMD2021}.

\section{Neural Networks, Deep Learning and Bayesian DL} 
The main objective of the NN for a regression problem can be described as follows:
\[
\xb_i \ra\fbox{NN: $\red{\phi}(\xb_i)$}\ra y_i
\]
Inferring an unknown function \(\red{f} : \mathbb{R}^M \to \mathbb{R}\) given the observations \(\yb = ( y_1, \dots,  y_N)^\text{T}\) at locations given by \(\xb = (\xb_1, \dots, \xb_N)\).  
The usual NN Learning approach is to define a parametric family of functions 
\(\phi_{\rthetab}:\mathbb{R}^M \times \rthetab \to \mathbb{R}\) flexible enough so that \(\exists \rthetab^\star\) such that \(\red{\phi}(\cdot) \approx \phi_{\rthetab^\star}(\cdot)\): 
\[
\{\xb_i,y_i\} \ra\fbox{NN Learning: $y_i=\red{\phi}(\xb_i) \approx \phi_{\rthetab^\star}(\xb_i)$}\ra \rthetab^\star
\]
Deep learning focuses on learning the optimal parameters \(\rthetab^{\star}\) which can then be used for predicting the output $\hat{y}_j$ for any input $\xb_j$: 
\[
\xb_j \ra\fbox{NN: $\phi_{\rthetab^\star}(\xb_j)$}\ra \hat{y}_j
\]
In this approach there is no uncertainty quantification. 

The Bayesian Deep learning can be summarized as follows:
\[
\{\xb_i,y_i\} \ra\fbox{$p(\rthetab|\Dc)=\frac{p(\Dc|\rthetab) p(\rthetab)}{p(\Dc)}$}\ra p(\rthetab|\Dc), 
\]
\[
\xb_j \ra\fbox{$\disp{p(y_j|\xb_j,\Dc)=\int p(y_j|\xb_j,\rthetab, \Dc) p(\rthetab|\Dc) \d{\rthetab}}$}\ra \hat{y}_j
\]
As we can see, uncertainties are accounted for both steps of parameter estimation and for prediction. However, as we will see the computational costs are important. We need to find solutions to do fast computation.

\section{Bayesian inference and approximate computation}
In a general Bayesian framework for NN and DL, the objective is to infer on the parameters $\rthetab$ from the data : $\Dc:\{\xb_i,y_i\}$, using the Bayes rule: 
\beq
p(\rthetab|\Dc)=\frac{p(\Dc|\rthetab) p(\rthetab)}{p(\Dc)},
\eeq
where $p(\rthetab)$ is the prior, $\ell(\Dc|\rthetab)=p(\Dc|\rthetab)$ is the likelihood, 
$p(\rthetab|\Dc)$ is the posterior, and \\ 
$\disp{p(\Dc)=\int p(\Dc|\rthetab) p(\rthetab) \d{\rthetab}}$ is called the Evidence. We can also write: 
\(
p(\rthetab|\Dc)\propto \ell(\Dc|\rthetab) p(\rthetab)
\), 
where the classical Maximum Likelihood estimation (MLE) is defined as: 
\( 
\rthetabh = \argmax{\rthetab}{\ell(\Dc|\rthetab)}
\). 
A particular point of the posterior is of high interest, because we may be interested to Maximum A Posterior (MAP) estimation: 
\(
\rthetabh = \argmax{\rthetab}{\ell(\Dc|\rthetab) p(\rthetab)}
\). 
We may also be interested in Mean Squared Error (MSE) estimation which is shown that it correspond to: 
\(
\disp{\rthetabh = \espx{p(\rthetab|\Dc)}{\rthetab} = \intg \rthetab \, p(\rthetab|\Dc) \d{\rthetab}}
\). 

The exact expression of the posterior and the computations of these integrals for great dimensional problems may be very computationally costly. For this reason, we need to do  Approximate computation. In the following subsections, we review a few solutions. 

\subsection{Laplace Approximation}
Rewriting the general Bayes rule slightly differently: 
\beq
p(\rthetab|\Dc)=\frac{1}{p(\Dc)} {p(\Dc|\rthetab) p(\rthetab)}=\frac{1}{Z} \expf{\Lc(\rthetab|\Dc)},
\eeq
the Laplace approximations use a second-order expansion of 
$\Lc(\rthetab|\Dc)=\ln p(\Dc|\rthetab)+\ln p(\rthetab)$ around $\rthetabh_{MAP}$ to construct a Gaussian approximation to $p(\rthetab|\Dc)$:
\beq
\Lc(\rthetab) \approx \Lc(\rthetabh_{MAP}) + \frac{1}{2} (\rthetab - \rthetabh_{MAP})' \left( \nabla^2 _\rtheta \Lc(\rthetab) |_{\rthetabh_{MAP}} \right)(\rthetab - \rthetabh_{MAP}),
\eeq
where the first-order term vanishes at $\rthetabh_{MAP}$. This is equivalent to do the following approximation: 

\bcc 
\fbox{~~Approximate~~ $p(\rthetab|\Dc)$ ~~by~~ $q(\rthetab)=\Nc\left(\rthetab|\rthetabh_{MAP},\Sigmab=[\nabla^2 _\rtheta \Lc(\rthetab) |_{\rthetabh_{MAP}}]\right)$~~} 
\ecc
With this approximation, the Evidence $p(\Dc)$ is approximated by:
\beq
Z=p(\Dc)= (2\pi)^{d/2} |\Sigmab|^{1/2} \expf{-\Lc(\rthetabh_{MAP})}.
\eeq

\REM{
Performing the Laplace approximation is illustrated in Fig~\ref{fig01}.
\begin{figure}[htb]
\bcc
\includegraphics[scale=1.2]{MAP_Laplace}\\
{\tt\small adapted from https://arxiv.org/abs/2106.14806}
\ecc
\caption{Laplace approximation: First find the MAP solution by any optimization method (left), then approximate the posterior $p(\rthetab|\Dc)$ by a Gaussian $q(\rthetab)=\Nc\left(\rthetab|\rthetabh_{MAP},\Sigmab=[\nabla^2 _\rtheta \Lc(\rthetab) |_{\rthetabh_{MAP}}]\right)$ around it (right). Figure adapted from\cite{1021167}}
\label{fig00}
\end{figure}
}

For great dimensional problems such as BDL, the full computation of $\Sigmab$ is very costly. We have still do more approximations. The following are a few solutions for scalable approximations for BDL :
\bit
\item Work with sub network or last layer (Transfer learning);
\item Covariance matrix decomposition (Low Rank, Kronecker-factored approximate curvature (KFAC), Diag);
\item Do the computation during the hyper parameter tuning using Cross validation;
\item Use for Approximate predictive computation
\eit

\subsection{Approximate computation: Variational inference}

The main idea then is to do Approximate Bayesian Computation (ABC) by approximating the posterior $p(\rthetab|\Dc)$ by a simpler expression $q(\rthetab)$. 
When approximation is done by minimizing 
\beq
\KL{q(\rthetab)}{p(\rthetab|\Dc)}=\intg q(\rthetab) \log\frac{q(\rthetab)}{p(\rthetab|\Dc)} \d{\rthetab}, 
\eeq
the method is called Variational Bayesian Approximation \emph{(VBA)}. 
When $q(\rthetab)$ is chosen to be separable in some components of the parameters:
\( 
q(\rthetab)=\prod_j q_j(\rtheta_j)
\), 
the approximation is called \emph{Mean Field VBA (MFVBA)}. 

Let come back to the general VBA:
\(
\disp{
\KL{q}{p}=\intg q \log\frac{q}{p} =\espx{q}{\log q}-\espx{q}{\log p}
}
\), 
and note by: 
$L =  \espx{q}{\log p}$, the expected likelihood and by  
$S =-\espx{q}{\log q}$, the entropy of $q$. Then, we have:
\beq
q*=\argmin{q}{\KL{q}{p}}=\argmax{q}{E=S+L}.
\eeq
$E$ is also called the Evidence lower bound (ELBO):
\beq
ELBO = -\espx{q}{\log q} + \espx{q}{\log p}.
\eeq
At this point, it is important to note one main property of VBA: When $p(\rthetab|\Dc)$, the posterior probability law $p$ and the approximate probability law $q$ are in the exponential family, then: 
$\espx{p}{\rthetab}=\espx{q}{\rthetab}$. 

\subsection{VBA and Natural Exponential family}
If $q$ is choosed to be in an \emph{natural exponential family}: 
\(
q(\rthetab|\etab)=\expf{\etab'\rthetab-A(\etab)}
\), 
then it is entirely characterized by its mean $\mb=\espx{q}{\rthetab}$,  
and if $q$ is conjugate to $p$, then:
\(
q^*(\rthetab|\etab)=
\expf{{\etab^*}'\rthetab-A(\etab^*)}
\), 
which is entirely characterized by its mean 
$\mb^*=\espx{q^*}{\rthetab}$.
We can then define the objective $E$ as a function of $\mb$ and the first order condition of the optimality is: 
\(
\dpdx{E}{\mb} \big|_{\mb=\mb*}=0
\). 
From this property, we can obtain a fixed-point algorithm to compute $\mb^*=\espx{q^*}{\rthetab}$:
\[
\dpdx{E}{\mb} \big|_{\mb=\mb*}=0 \ra 
\dpdx{E}{\mb} \big|_{\mb=\mb*}+\mb=\mb \ra \Mb(\mb)=\mb. 
\]
Iterating on this \emph{fixed point algorithm}:
\beq
\Mb(\mb)=\mb^{(k-1)} \quad \mbox{with}\quad 
\Mb(\mb):=\dpdx{E}{\mb}+\mb
\eeq
converges to $\mb^*=\espx{q^*}{\rthetab^*}$ which is also $\mb^*=\espx{p}{\rthetab^*}$.  
This algorithm can be summarized as follows: 
\bit
\item Having chosen prior and likelihood, find the expression of $p(\rthetab|\Dc)\propto p(\Dc|\rthetab) p(\rthetab)$;
\item Choose a family $q$ and find the expressions of $L=\espx{q}{\ln p}$ and  $S=\espx{q}{\ln q}$, and thus $E=L+S$ as a function of $\mb=\espx{q}{\rthetab}$;
\item Find the expression of the vector operator $\Mb=\nabla_{\mb} E + \mb$ and update it $\Mb(\mb)=\mb^{(k-1)}$ until convergence which results to  
$\mb^*=\espx{q^*}{\rthetab^*}=\espx{p^*}{\rthetab^*}$. 
\eit

At this point, it is important to note that, in this approach, even if the mean is well approximated, the variances or the covariances are under estimated. Some authors interested in this approach have proposed solutions to better estimate the covariances.  
See 
\cite{Giordano2015} 
for one of the solutions. 

\section{Deep learning and Bayesian Bayesian DL} 
As introduced before, in classical DL, the training and prediction steps can be summarized as follows: 
\[
\{\xb_i,y_i\} \ra\fbox{NN Learning: $\phi_{\rthetab^\star}(\cdot)$}\ra \rthetab^\star
\qquad 
\xb_j \ra\fbox{NN: $\phi_{\rthetab^\star}(\xb_j)$}\ra \hat{y}_j
\]
In this approach there is no uncertainty quantification. 
The Bayesian Deep learning can be summarized as follows:
\[
\{\xb_i,y_i\} \ra\fbox{$p(\rthetab|\Dc)=\frac{p(\Dc|\rthetab) p(\rthetab)}{p(\Dc)}$}\ra p(\rthetab|\Dc) \mbox{~~or~~} q(\rthetab|\Dc),
\]
\[
\xb_j \ra\fbox{$\disp{p(y_j|\xb_j,\Dc)=\int p(y_j|\xb_j,\rthetab, \Dc) p(\rthetab|\Dc) \d{\rthetab}}$}\ra p(y_i|\xb_j,\Dc) \mbox{~~or~~} q(y_i|\xb_j,\Dc).
\]
As we can see, uncertainties are accounted for both steps of parameter estimation and for prediction. However, computational costs are important. We need to find solutions to do fast computation. As mentioned before, the different possible tools are: Laplace and Gaussian approximation, Variational Inference, and more controlled approximation to design new deep-learning algorithms which can scale up for practical situations. 

\subsection{Exponential family approximation} 
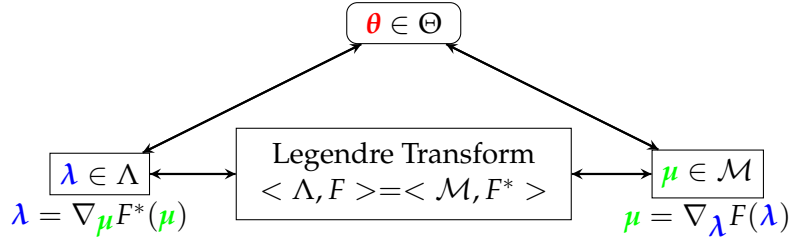
\begin{figure}[htb]
\bcc
\begin{tikzpicture}[node distance=1cm]
\node (Orig) [startstop] {$\rthetab\in\Theta$};
\node (LT) [process, below of=Orig, yshift=-1cm] {$\barr{c} \mbox{Legendre Transform} \\ <\Lambda, F> = <\Mc, F^*> \earr$};
\node (Nat) [process, below of=Orig, yshift=-1cm, xshift=-4cm] {$\blue{\lambdab}\in\Lambda$};
\node (Exp) [process, below of=Orig,  yshift=-1cm, xshift=+4cm] {$\green{\mub}\in\Mc$};
\draw [arrow](Orig) -- (Exp);
\draw [arrow](Exp) -- (Orig);
\draw [arrow](Orig) -- (Nat);
\draw [arrow](Nat) -- (Orig);
\draw [arrow](Nat) -- (LT);
\draw [arrow](LT) -- (Nat);
\draw [arrow](Exp) -- (LT);
\draw [arrow](LT) -- (Exp);
\node (pf) [note, below of=Nat,yshift=4mm] {$\blue{\lambdab}=\nabla_{\green{\mub}} F^*(\green{\mub})$};
\node (pf) [note, below of=Exp,yshift=4mm] {$\green{\mub}=\nabla_{\blue{\lambdab}} F(\blue{\lambdab})$};
\end{tikzpicture}
\ecc
\vspace{-12pt}
\caption{General Exponential family, with Original parameters $\rthetab$, Natural parameters $\blue{\lambdab}$, Expectations parameters $\green{\mub}:= \espx{q}{\tb(\rthetab)}$ and their relations via the Legendre Transform.}
\label{fig01}
\end{figure}

Let consider the case of general exponential families 
$q(\rthetab|\emph{\lambdab})=\expf{\blue{\lambdab}^T \tb(\rthetab)-F(\blue{\lambdab})}$, where: 
$\rthetab$ represents the original parameters,  
$\blue{\lambdab}$ represents the natural parameters, 
$\tb(\rthetab)$ sufficient statistics, $F(\blue{\lambdab})$ log partition function and define the Expectations parameters $\green{\mub}:= \espx{q}{\tb(\rthetab)}$. 
Let also define the Dual function $F^*(\etab)$ and dual parameters $\etab$ via the 
Legendre transform:
\[
G(\etab)=F^*(\etab)=\sup_{\lambdab}\{<\lambdab, \etab>-F(\lambdab)\}.
\]
Then, we can show the triangular relation between $\rthetab\in\Theta$, $\blue{\lambdab}\in\Lambda$ and $\green{\mub}\in\Mc$ in Fig~\ref{fig01}. 

With these notations, applying the VBA rule: \quad 
\(
\min_{q\in \Qc} \espx{q}{\Lc(\rthetab)} - H(q)
\), \; 
results to the following updating rule for natural parameters:
\beq
\lambdab \la \lambdab -\rho \nabla_{\mub}
\left(\Lc(\rthetab) - H(q)\right).
\eeq
For example, considering the Gaussian case: 

\beq
q(\rthetab)=\Nc(\rthetab|\mb,\Sb^{-1})
\propto \expf{-\frac{1}{2}(\rthetab-\mb)^T\Sb(\rthetab-\mb)} 
\propto \expf{(\Sb\mb)^T\rthetab+\mbox{Tr}(-\frac{\Sb}{2}\rthetab\rthetab^T)}. 
\eeq
we can identify the natural parameters as: 
\blue{$\lambdab= [\Sb\mb, -\Sb/2]\ra 
\mub:= [\espx{q}{\rthetab}, \espx{q}{\rthetab\rthetab^T}]$}, 
and we get easily the following algorithm: 
\beq
\left\{\barr{l}
\Sb\mb \la (1-\rho) \Sb\mb - \rho \nabla_{\espx{q}{\rthetab}} \espx{q}{\Lc(\rthetab)},
\\ 
\Sb\la (1-\rho) \Sb - \frac{\rho}{2} \nabla_{\espx{q}{\rthetab\rthetab^T}} \espx{q}{\Lc(\rthetab)}.
\earr\right.
\eeq
where 
\[
\left\{\barr{l}
\nabla_{\espx{q}{\rthetab}} \espx{q}{\Lc(\rthetab)}=
\espx{q}{\nabla_{\rthetab} \Lc(\rthetab)}
-2\espx{q}{\Lc(\rthetab)},
\\
\nabla_{\espx{q}{\rthetab\rthetab^T}} \espx{q}{\Lc(\rthetab)}=
\espx{q}{\Hb(\rthetab)},
\earr\right.
\]
which results, explicitly, to:
\beq
\left\{\barr{l}
\mb \la \mb-\rho \Sb^{-1} \nabla_\mb {\Lc(\mb)},
\\
\Sb \la (1-\rho) \Sb + \rho \Hb_\mb.
\earr\right.
\eeq
For a linear model: $\yb=\Xb\rthetab$ and Gaussian priors, we have:
\beq
\Lc(\rthetab)
=(\yb-\Xb\rthetab)^T(\yb-\Xb\rthetab)+\gamma\rthetab^T\rthetab
=-2\rthetab^T(\Xb^T\yb)+\Tr{\rthetab\rthetab^T (\Xb^T\Xb+\gamma\Ib)}, 
\eeq
and $\espx{q}{\Lc(\rthetab)}=\lambdab^T \mub, \quad 
\nabla_{\mub} \espx{q}{\Lc(\rthetab)}=\lambdab $.

It is interesting to note that many classical algorithms for updating the parameters such as Forward-Backward, Sparse Variational Inference and Variational Message Passing become special cases. 
A main remark here is that, this linear generating function case, will link us to the linear inverse problems $\gb=\Hb\rfb+\epsilonb$ if we replace $\yb$ by $\gb$, $\Xb$ by 
$\Hb$ and $\rthetab$ by $\rfb$. 

\section{Linear inverse problems and BDL}
Linear inverse problems $\bgb=\Hb\rfb+\epsilonb$ where we know the forward model $\Hb$ and we assume Gaussian likelihood $p(\gb|\rfb)=\Nc(\bgb|\Hb\rfb,\sigmae^2\Ib)$ and Gaussian prior $p(\rfb)=\Nc(\rfb|\zerob,\sigma_f^2\Ib)$ is the easiest case to consider, where, we know that the posterior is Gaussian $p(\rfb|\bgb)=\Nc(\rfb|\rfbh,\Sigmabh)$ with 
\[
\rfbh=(\Hb^t\Hb+\lambda \Ib)^{-1}\Hb^t \bgb = \Ab \, \bgb =\Bb \Hb^t \, \bgb
\mbox{~~or still~~} 
\rfbh=\Hb^t(\frac{1}{\lambda}\Hb\Hb^t+\Ib)^{-1}\bgb =\Hb^t \Cb \, \bgb, 
\mbox{~with~} \lambda=\sigmae^2/\sigma_f^2, 
\]
where $\Ab=(\Hb^t\Hb+\lambda \Ib)^{-1}\Hb^t$, $\Bb=(\Hb^t\Hb+\lambda \Ib)^{-1}$ and $\Cb=(\frac{1}{\lambda}\Hb\Hb^t+\Ib)^{-1}$ and $\Sigmabh=(\Hb^t\Hb+\lambda \Ib)^{-1}$. 

These relations can be presented schematically as 
\[
\bgb\ra\fbox{$~~~~~~\Ab~~~~~~$}\ra\rfbh, \qquad 
\bgb\ra\fbox{$~~\Hb^t~~$}\ra\fbox{$~~\Bb~~$ }\ra\rfbh, \qquad 
\bgb\ra\fbox{$~~\Cb~~$}\ra\fbox{$~~\Hb^t~~$ }\ra\rfbh
\]
We can then consider to replace $\Ab$, $\Bb$, and $\Cb$ by appropriate Deep Neural networks and apply all the previous BDL methods to them. 

We can also, consider unfolding technics of the MAP or VBA optimization algorithms as the basic structure of the DLNN and then apply the BDL parameter estimation methods such as the following case taken from  
\cite{review2017mccann,one2017chang,deep2020liang,review2017unser,physicsinformed2019chen}. 

\def\SoftT{
\bpic(25,20)
  \put(0,10){\line(1,0){20}}
  \put(10,0){\line(0,1){20}}
  \put(15,10){\line(1,1){10}}
  \put(-5,0){\line(1,1){10}}
\epic
}

\def\Blockwx#1#2{
\bpic(150,80)
  \put(20,20){\vector(1,0){10}}
  \put(30,10){\framebox(30,20){#2}}
  \put(60,20){\vector(1,0){10}}
  \put(80,20){\circle{20}}
  \put(80,20){\makebox(0,0){$+$}}
  \put(90,20){\vector(1,0){10}}
  \put(100,10){\framebox(30,20){\SoftT}}
  \put(130,20){\vector(1,0){10}}
  \put(80,40){\vector(0,-1){10}}
  \put(70,40){\framebox(20,15){#1}}
  \put(80,67){\vector(0,-1){10}}
  \put(75,70){\makebox(10,10){$\bgb$}}
  \put(80,75){\circle{20}}
\epic
}
\def\BlockDLw{\begin{picture}(100,100)
  \put(-160,10){\Blockwx{$\rWb_0$}{$\rWb^{(1)}$}}
  \put(-50,10){\Blockwx{$\rWb_0$}{$\rWb^{(2)}$}}
  \put(90,10){\Blockwx{$\rWb_0$}{$\rWb^{(K)}$}}
  \put(103,30){\makebox(0,0){......}}
  \put(-140,35){\makebox(0,0){$\rfbh^{(1)}$}}
  \put(245,35){\makebox(0,0){$\rfbh^{(K)}$}}
\end{picture}
}

\begin{figure}[htb!]
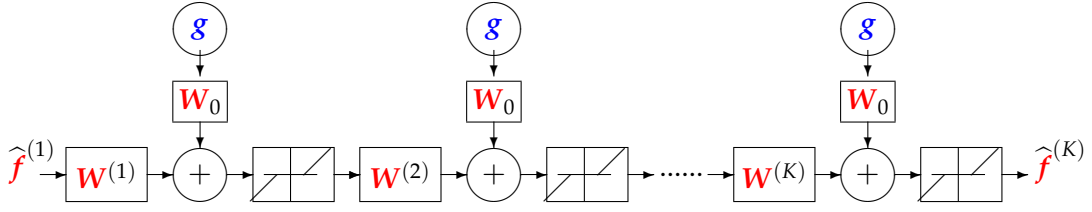

\bcc
\BlockDLw
\ecc
\vspace{-18pt}
\caption{The $K$ layers of the DNN are equivalent to $K$ iterations of an iterative gradient based optimization algorithm. The simplest solution is to choose 
$\rWb_0=\alpha\Hb$ and  
$\rWb^{(k)}=\rWb=(\Ib-\alpha\Hb^t\Hb), \quad k=1,\cdots,K$, thus learning the only parameters $\rWb_0$ and $\rWb$. 
A more robust, but more costly is to learn all the layers $\rWb_0=\alpha\Hb$ and 
$\rWb^{(k)}=(\Ib-\alpha^{(k)}\Hb^t\Hb), \quad k=1,\cdots,K$.}
\label{fig02}
\end{figure}

\section{DL structure and deterministic or Bayesian computation}

To be able to look at the DNN and analyse it either in a deterministic or Bayesian, 
let come back to the general notations and consider the following NN with input $\xb$, output $\yb$ and intermediate hidden variables $\zb_1, \zb_2,...,\zb_L$: 
\[
\xb \ra\fbox{first layer}\ra \zb_1\ra\fbox{second layer}\ra\zb_2\cdots \ra\fbox{last layer}\ra \yb.
\]
In the deterministic case, each layer is defined by its parameters $(W_l,b_l)$: 
\[
\xb \ra\fbox{$f_{W_0,b_0}(\xb)$}\ra \zb_1\ra\fbox{$f_{W_1,b_1}(\zb_1)$}\ra\zb_2\cdots \ra\fbox{$f_{W_L,b_L}(\zb_{L-1})$}\ra \yb, 
\]
and we can write: \\ 
\(
\yb=f_{\thetab}(\xb)=(f_{(W_0,b_0)} \odot f_{(W_1,b_1)} \odot \cdots f_{(W_L,b_L)})(\xb)
\)  
with 
\(
\rthetab=({(W_0,b_0)},{(W_1,b_1)},\cdots,{(W_L,b_L)})
\)
and 
\beq
\yb  =f_L(W_L\zb_{L-1}+b_L), \quad 
\zb_l=f_{l}(W_l\zb_{l-1}+b_l), \quad l=L,\cdots,1, \quad   
\zb_{0}=\xb.
\eeq

\subsection{Deterministic DL computation} 
In general, during the training steps, the parameters $\rthetab$ are estimated via: 
\beq
\rthetab^*=\argmin{\rthetab}{\sum_i \ell(y_i,f_{\rthetab}(\xb_i))+\sum_k \blue{\lambda_k} \phi_k(\red{W_k,b_k})}.
\eeq
The main point here is to choose how to choose $\lambda_k$ and $\phi(.)$ and which optimization algorithm to choose for better convergence. 

When parameters are obtained (the model is trained), we can use it easily via: 
\beq
\zb_{0}=\xb, \quad  
\zb_l=f_{l}(W_l\zb_{l-1}+b_l), \quad   
\yb  =f_L(W_L\zb_{L-1}+b_L).
\eeq

\subsection{Bayesian Deep Neural Network} 
In Bayesian DL, the question of choosing $\lambda_k$ and $\phi(.)$ in the previous case becomes the choice of the prior $p(\rthetab)$ which can also be assumed a priori separable in the components of $\rthetab$ or not. Then, we have to choose the expression of the likelihood (in general Gaussian) and find the expression of the posterior $p(\rthetab|\Dc)$. As explained extensively before, directly using this posterior is almost impossible. Hopefully, we have a great number of approximate computation methods, such as the MCMC sampling, slice sampling, nested sampling, data augmentation and  variational inference which can still be used in practical situations. However, the training step in the Bayesian DL still stays very costly, in particular, if we want to quantify uncertainties. 

\REM{
\subsubsection{Data Augmentation methods}
There are many works, proposing approximate solutions \cite{}. Between all, we can mention data augmentation methods, where the main idea is to introduce a vector of auxiliary variables, 
$\omegab_1, \cdots, \omegab_i$ such that: 
\(
p(\thetab|\Dc)=\espx{p(\omegab)}{p(\omegab,\thetab|\Dc}
\), 
and 
\(
p(\omegab,\thetab|\Dc)=p(\thetab|\omegab,\Dc) \, p(\omegab|\Dc)
\). 
Then, $p(\thetab|\omegab,\Dc)$ can be approximated by a conditional Gaussian and we can use what is called \emph{Data Augmentation trick}: 
\[
\expf{\ell(\yb,f_{\thetab}(\xb))}=\espx{p(\omegab)}{-Q(\yb|f_{\thetab}(\xb),\omegab)} 
\]
Now, the problem is how to choose $p(\omegab)$ in such a way that for different activation function, we obtain a quadratic function for $Q$ ? For some specific activation functions, there are analytical solutions, for example those shown here: 
\\
{\small \[
\barr{ll|l}
\mbox{activation function} & & p(\omega) \\ \hline
\mbox{ReLU}: 
\exp(-max(1-x,0))
&=\espx{p(\omega)}{\Nc(x|(1+\omega^2, \omega)} & \Gc\Ic\Gc(1,0,0)  
\\
\mbox{Logit}: \exp(-(1+e^x))
&=\espx{p(\omega)}{\Nc(x|(0, 1/\omega)} & \Pc\Gc(1,0) \\
\mbox{Lasso}: \exp(-|x|)
&=\espx{p(\omega)}{\Nc(x|(0, 1)} & \Ec(1/2)  \\
\earr
\]
}
\\ 
where 
$\Gc\Ic\Gc$ is the Gaussian Inverse Gaussian,  \; 
$\Pc\Gc$ is the Polya Gamma, and  
$\Ec$: is the Exponential distributions. 
}

\subsubsection{Prediction step}
In the prediction step also, again, we have to consider choosing a probability law $p(\xb)$ for the class of the possible inputs, and for all the outputs $\zb_l$ conditional to their inputs $\zb_{l-1}$. Then, we can consider, for example the Gibbs sampling scheme. A comparison between deterministic and Bayesian DL is shown here:  
\[
\barr{lcl}
\mbox{Deterministic:} & & \mbox{Bayesian:}
\\
\left\{\barr{l}
\zb_{0}=\xb, \\ 
\zb_l=f_{l}(W_l\zb_{l-1}+b_l), \quad l=1,\cdots,L,\\  
\yb  =f_L(W_L\zb_{L-1}+b_L).
\earr\right.
&\ra&
\left\{\barr{l}
\zb_{0}\sim p(\xb),\\ 
\zb_l\sim p(\zb_l|\zb_{l-1}), \quad l=1,\cdots,L,\\ 
\yb  \sim p(\yb|\zb_{L-1}).
\earr\right.
\earr
\]
If we consider Gaussian laws for the input and all the conditional variables, then we can write: 
\beq
\zb_{0}\sim N(\zb_0|\xb,\tau_0\Ib), \qquad 
\zb_l\sim N(\zb_l|f_l(\zb_{l-1}),\tau_l\Ib), \; l=1,\cdots,L, \qquad  
\yb  \sim p(\yb|\zb_{L-1}). 
\eeq
Here too, the main difficulty occurs when there are non-linear activation functions, in particular in the last layer, where the Gaussian approximation may no more be valid.


\section{Application: Infrared imaging}
Infrared (IR) imaging is used to diagnosis and to survey the temperature field distribution of sensitive objects in many industrial applications. These images are, in general, low resolution and very noisy. The real values of the temperature depends also to many other parameters, such as emissivity, attenuation and diffusion due to the distance of the camera to the sources, etc. To be really useful in practice, we need to reduce the noise, to calibrate, to increase the resolution, to segment for detection the hot area, and finally survey the temperature values of the different area during the time to be able to do preventive diagnosis and possibly maintenance. 

Reducing the noise can be done by filtering using Fourier Transform or Wavelet Transform or other sparse representation of images. To increase the resolution, we may use deconvolution methods if we can get the point spread function (PSF) of the camera, or if not by blind deconvolution technics. The segmentation and detection of the hot area and temperature value estimation at each area is also very important step in real applications. 
Any of these steps can be done separately, but trying to propose a global processing using DL or BDL is necessary for real applications. As, any of these steps, are in fact different inverse problems, and it is difficult to fix the parameters in each step in a robust way, we propose a global process using the BDL. 

\begin{figure}[hbt]
\[
\barr{@{}c@{}c@{}c@{}l@{}c@{}}
\mbox{Input~} \bgb & denoising & deconvolution & Segmentation& \mbox{Segmented}\\
\mbox{IR image}&\ra\fbox{\fbox{\fbox{C1}-\fbox{Th}-\fbox{C2}}}\ra &
\fbox{\fbox{\fbox{C3}-\fbox{Thr}-\fbox{C4}}}\ra
&\fbox{\fbox{~~Bayesian SegNet~~}}\ra &
\mbox{image}
\earr
\]
\caption{The proposed 4 groups of layers NN for denoising, deconvolution and segmentation of IR images.}
\label{fig03}
\end{figure}

\vspace{-1.2cm}
\begin{figure}[hbt]
\includegraphics[height=6cm,width=18cm]{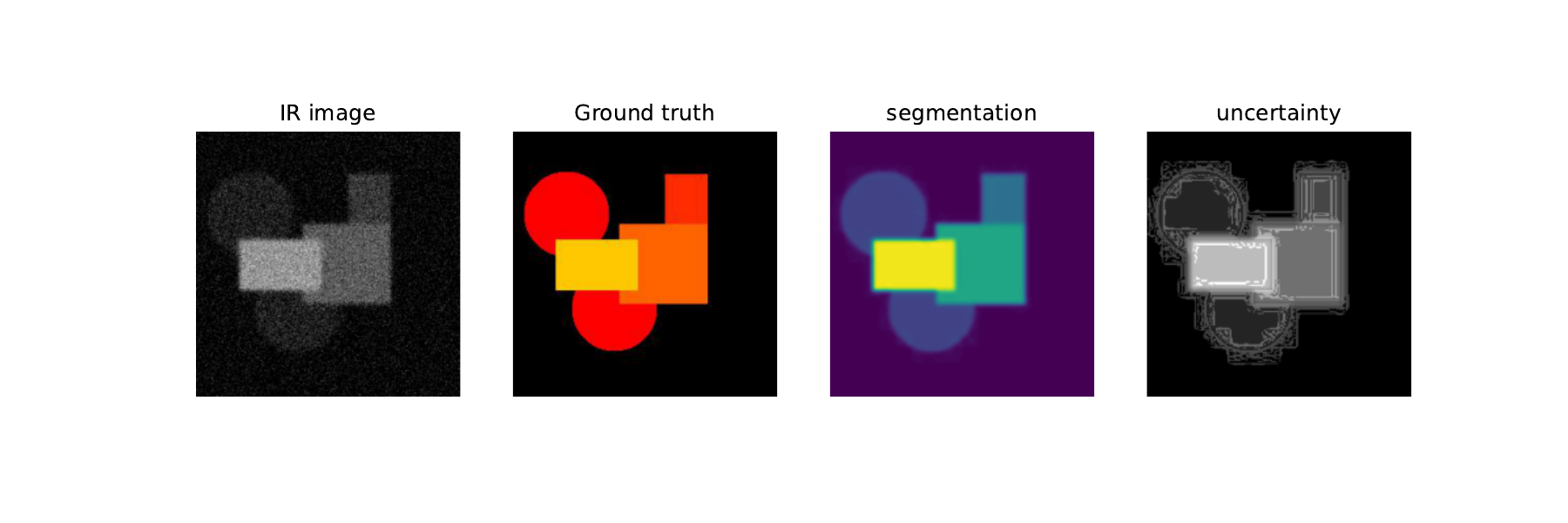}

\vspace{-1.5cm}
\includegraphics[height=6cm,width=18cm]{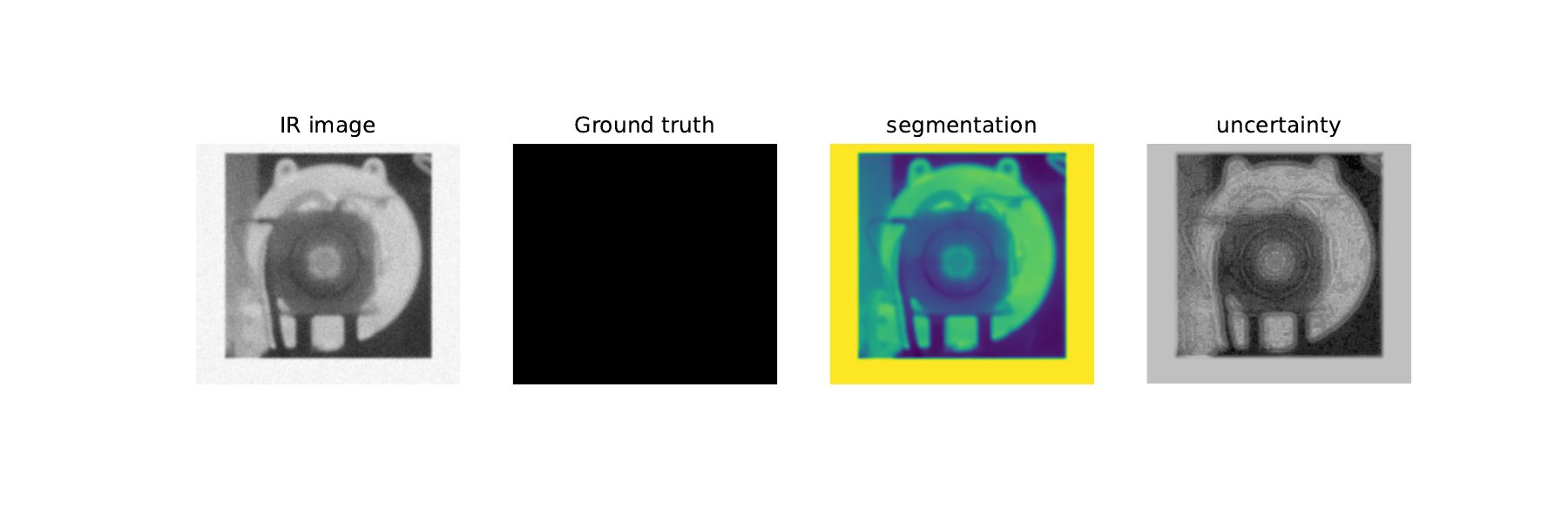}

\vspace{-1.5cm}
\caption{Example of expected results: First row from left: a) simulated IR image, b) its ground truth labels, c) the result of the deconvolution and segmentation and d) uncertainties. Second row: e) a real IR image, f) no ground truth, g) the result of its deconvolution and segmentation, and h) uncertainties.}
\label{fig04}
\end{figure}

In a first step, as the final objective is to segment image to obtain different levels of temperature (for example 4 levels:backgrounds, normal, high, and very high), we propose to design a NN which gets as input a low resolution and noisy image and outputs a segmented image with those 4 levels and at the same time a good estimate of the temperature at each segment. 

To train this NN, we can generate different known shaped images to consider as the ground truth and simulate the blurring effects of temperature diffusion  via the convolution of different appropriate PSF and add some noise to generate realistic images. We can also use a black body thermal sources and acquire different images at different conditions. All these images can be used for the training of the network. 

\def\url#1{\tt#1}

\bibliographystyle{Inputs/ieeetr.bst}
\bibliography{biblio/IEEEXplore2020,biblio/IP_ML_IEEE,biblio/amd_2022,biblio/litmaps_ML_IP}

\end{document}